\documentclass[nohyperref]{article}

\usepackage{microtype}
\usepackage{graphicx}
\usepackage{subfigure}
\usepackage{booktabs} 

\usepackage{hyperref}
\usepackage{algorithm2e}
\usepackage{lscape}
\usepackage{xcolor}

\usepackage[accepted]{icml2022}

\usepackage{amsmath}
\usepackage{amssymb}
\usepackage{mathtools}
\usepackage{amsthm}
\usepackage[capitalize,noabbrev]{cleveref}
\usepackage{float}

\usepackage{amsmath,amsfonts,bm}

\def\eqref#1{equation~\ref{#1}}

\def\1{\bm{1}}

\DeclareMathAlphabet{\mathsfit}{\encodingdefault}{\sfdefault}{m}{sl}
\SetMathAlphabet{\mathsfit}{bold}{\encodingdefault}{\sfdefault}{bx}{n}

\newcommand{\E}{\mathbb{E}}

\DeclareMathOperator*{\argmax}{arg\,max}

\theoremstyle{plain}

\theoremstyle{definition}

\theoremstyle{remark}

\DeclareMathOperator*{\mean}{mean}
\DeclareMathOperator*{\std}{std}

\usepackage[textsize=tiny]{todonotes}

\icmltitlerunning{Reducing Variance in Temporal-Difference Value Estimation via Ensemble of Deep Networks}

\begin{document}

\twocolumn[
\icmltitle{Reducing Variance in Temporal-Difference Value Estimation \\ via Ensemble of Deep Networks}
\begin{icmlauthorlist}
    \icmlauthor{Litian Liang}{UCI}
    \icmlauthor{Yaosheng Xu}{UCI}
    \icmlauthor{Stephen McAleer}{CMU}
    \icmlauthor{Dailin Hu}{UCI} \\
    \icmlauthor{Alexander Ihler}{UCI}
    \icmlauthor{Pieter Abbeel}{UCB}
    \icmlauthor{Roy Fox}{UCI}
\end{icmlauthorlist}
\icmlaffiliation{UCI}{University of California, Irvine}
\icmlaffiliation{CMU}{Carnegie Mellon University}
\icmlaffiliation{UCB}{University of California, Berkeley}
\icmlcorrespondingauthor{Litian Liang and Roy Fox}{\{litianl1, royf\}@uci.edu}
\icmlkeywords{
    Machine Learning, 
    Deep Learning, 
    Reinforcement Learning, 
    Deep Reinforcement Learning, 
    Off-Policy Reinforcement Learning, 
    Target Network, 
    Variance Reduction}
\vskip 0.3in
]



\printAffiliationsAndNotice{}

\begin{abstract}

In temporal-difference reinforcement learning algorithms, variance in value estimation can cause instability and overestimation of the maximal target value. Many algorithms have been proposed to reduce overestimation, including several recent ensemble methods, however none have shown success in sample-efficient learning through addressing estimation variance as the root cause of overestimation. In this paper, we propose MeanQ, a simple ensemble method that estimates target values as ensemble means. Despite its simplicity, MeanQ shows remarkable sample efficiency in experiments on the Atari Learning Environment benchmark. Importantly, we find that an ensemble of size 5 sufficiently reduces estimation variance to obviate the lagging target network, eliminating it as a source of bias and further gaining sample efficiency. We justify intuitively and empirically the design choices in MeanQ, including the necessity of independent experience sampling. On a set of 26 benchmark Atari environments, MeanQ outperforms all tested baselines, including the best available baseline, SUNRISE, at 100K interaction steps in 16/26 environments, and by 68\% on average. MeanQ also outperforms Rainbow DQN at 500K steps in 21/26 environments, and by 49\% on average, and achieves average human-level performance using 200K ($\pm$100K) interaction steps. Our implementation is available at \url{https://github.com/indylab/MeanQ}.

\end{abstract}

\section{Introduction}
Model-free reinforcement learning (RL) with deep learning has proven successful at achieving master-level performance in many sequential decision making problems \citep{alphago,alphazero,mnih2015human,starcraft2,kalashnikov2018qtopt,haarnoja2018algorithmapp}. RL algorithms learn a control policy that maximizes the expected discounted sum of future rewards (the \emph{policy value}) through experience collected via interacting with an environment. Temporal-Difference (TD) \citep{sutton2018reinforcement} is a principled approach to RL that maintains value estimates and iteratively improves them using \emph{bootstrapped} targets that combine experienced short-term rewards and current estimates of future long-term values. 

Q-learning \citep{watkins1992q} learns state–action value estimates (\emph{Q function}) by minimizing the TD error between the estimates and the bootstrapped targets. Although tabular Q-learning asymptotically converges to the optimal policy \citep{tsitsiklis1994asynchronous,jaakkola1994convergence}, it is known to be positively biased and overestimate the value before convergence due to Jensen's inequality under value estimation uncertainty~\citep{Thrun-1993-15908,fox2016taming}. This bias is detrimental to efficient learning, because it propagates through bootstrapping and because it can cause further experience collection to use suboptimal actions that appear optimal. 
Deep Q-Networks~(DQN) \citep{DBLP:journals/corr/MnihKSGAWR13,mnih2015human} represent the Q function with a deep neural network to learn expressive policies in environments with high dimensional states. 
Unfortunately, such value function approximation can further exacerbate instability and overestimation~\citep{Thrun-1993-15908,mahmood2015emphatic}.

One way to reduce the variance (between-runs variability) and instability (within-run variability) of value estimation 
is to use an ensemble of estimators. Ensemble learning is well-studied in machine learning and is known for its property of reducing estimation variance and ability to capture epistemic (model) uncertainty, usually in a form of prediction variance. In the field of RL, ensemble RL methods \citep{wiering2008ensemble} have been applied to improve exploration \citep{osband16a,osband16b,chen2017ucb,noisyExplore} and guide value or policy updates \citep{vanhasselt2015deep, fujimoto2018addressing, Fox2019Toward, lan2020maxmin, lee2021sunrise, Liang2021Temporal, chen2021randomized}. 

In this work, we propose a simple model-free ensemble method, called MeanQ, that reduces variance and instability of the TD target estimates by averaging an ensemble of neural network learners. We discuss similarities and differences between MeanQ and several closely related existing methods, including  \citet{DBLP:journals/corr/AnschelBS16}, and justify our specific design choices intuitively and empirically. We discuss the variance-reduction properties of MeanQ's target estimator and show empirically that, in some cases, instability is sufficiently reduced to eliminate the need for a target network~\citep{mnih2015human,lillicrap2015continuous,Kim2019DeepMellowRT}, which removed the target lag and further improves sample efficiency. 

In experiments in the Atari Learning Environment~(ALE) domain~\citep{bellemare2013arcade}, MeanQ significantly outperforms all baselines with which we compared, in a benchmark set of 26 environments. MeanQ outperforms the best available baseline, SUNRISE~\citep{lee2021sunrise}, at 100K interaction steps in 16/26 environments, and by 68\% in normalized return averaged over all 26 environments. Compared with a commonly used baseline, Rainbow DQN~\citep{hessel2018rainbow}, MeanQ achieves higher returns at 500K steps in 21/26 environments, and 49\% higher average normalized return. MeanQ also achieves average human-level performance in the 26 games using only 200K ($\pm$100K) interaction steps.

\section{Preliminaries}
We consider a Markov Decision Process (MDP) with probability $p(s'|s,a)$ to transition to state $s'$ when taking in state $s$ an action $a$ in a finite action space. An agent controls the dynamical process using a policy $\pi$ with probability $\pi(a|s)$ to take action $a$ in state $s$, after which a reward $r(s, a)$ is observed.

The MDP and policy jointly induce a distribution $p_\pi$ over the trajectory $\xi = (s_0, a_0, r_0, s_1 \ldots)$ in each \emph{episode}. An RL algorithm should discover a control policy that maximizes the expected discounted return, $R(\xi) = \sum_{t\ge0} \gamma^t r_t$, of trajectory $\xi$, where $t$ is the time step in the interaction process, $0 \le \gamma < 1$ is a discount factor, and $r_t = r(s_t, a_t)$. The value-to-go (Q~value) of policy $\pi$, starting at state–action pair $(s,a)$, is
$$Q_{\pi}(s,a) = \E_{\xi \sim p_\pi}[R(\xi)|s_0=s, a_0=a].$$
Value-based RL methods maintain a state–action value function (Q function) 
to infer a control policy and guide policy updates. Q-learning \citep{watkins1992q}
learns the optimal value of each state–action by stochastically updating a tabular representation of Q, on experience $(s, a, r, s')$, by
$$Q(s,a) \leftarrow Q(s,a) + \alpha(r + \gamma \max_{a'}Q(s',a') - Q(s,a)),$$
to minimize the Temporal-Difference (TD) error in parentheses.
Deep Q-Networks (DQN) \citep{DBLP:journals/corr/MnihKSGAWR13,mnih2015human} learn a parametrized $Q_\theta$ value function by minimizing the squared error between $Q_\theta$ and the target value estimate, typically predicted by a lagging version $Q_{\bar\theta}$ of the current value function
$$\mathcal{L}(s,a,r,s';\theta)=(r+ \gamma \max_{a'} Q_{\bar{\theta}}(s',a')-Q_\theta(s,a))^2.$$

The stochastic fitting process means that, before convergence of the algorithm, the Q values are uncertain value estimates.
The $\max$ operator in the noisy target value estimate has been shown to introduce an overestimation bias \citep{Thrun-1993-15908}, known as the “winner's curse”, due to Jensen's inequality
\begin{equation} \label{eq:jensen}
    \E[\max_{a'} Q(s',a')] - \max_{a'} \E[Q(s',a')] \ge 0,
\end{equation}
where the expectation is over runs of the algorithm.
Since the update is applied repeatedly through bootstrapping, it can iteratively increase the bias of the estimated Q values before convergence, and introduce instability into TD-learning algorithms as it propagates through the repeated Bellman update \citep{kumar2019stablizing,kumar2020discor} and neural-network extrapolation errors \citep{hasselt2018deep,lee2021sunrise}.

\section{Why Reduce Estimation Variance}\label{section_reduce_variance}

To take useful update steps in parameter space, a sample-efficient value-based RL algorithm must be able to produce informative next-state target estimates, even in early stages of training when data is scarce and estimates are noisy. Unfortunately, stochasticity in the fitting process, including parameter initialization, environment dynamics, exploration, and replay sampling, leads to randomness in the parameters and the target estimates. This uncertainty can introduce bias via the Jensen gap as well as destabilize the training process, particularly under function approximation. 

While the Jensen gap (\ref{eq:jensen}) cannot be fully characterized in terms of the variance of the value estimates, instead requiring extreme-value analysis, it is clear that target estimation variance generally contributes to overestimation bias in parametrized Q functions~\citep{Thrun-1993-15908,Kim2019DeepMellowRT,DBLP:journals/corr/abs-2001-02811}. Reducing the variance of the value estimate can help reduce the resulting bias. 

In the case of learning an approximate Q function, such as a neural network, variance in the target estimate (that is, estimate variability between runs) can also manifest as instability (that is, variability throughout a single run of the learning process).
One possible reason is that update steps with similar experiences, which should be similar, can become inconsistent in the presence of high target variance.
A target network \citep{mnih2015human,lillicrap2015continuous} regularizes the estimates by separating the learner's parameters from those responsible for producing target estimates. The target network uses a delayed copy or a Polyak-Rupert (exponential window) average of the parameters, effectively preventing it from changing too erratically.  

Target networks have been shown to successfully stabilize off-policy TD learning \citep{hasselt2018deep}. However, by introducing a lagging network for value estimation, target networks introduce additional bias to the target estimate in exchange for lower variance. In many algorithms, the target network update rate is a hyperparameter that requires careful tuning, and can be domain-dependent.

By averaging an ensemble of Q networks, MeanQ reduces the target estimate variance. Empirically, this reduces both the overestimation bias and the instability that this variance can cause. Thus MeanQ enjoys better sample efficiency, which is even further improved by using up-to-date target estimates, since a lagging target network is shown to be unneeded for stability in MeanQ.  

\section{Related Work}
\subsection{Off-Policy RL}
Off-policy RL algorithms improve sample efficiency by reusing environment interactions experienced by different policy \citep{fujimoto2018addressing,haarnoja2018soft,hessel2018rainbow}. Off-policy learning is thus considered a promising direction for scaling RL to meet the needs of the real world \citep{levine2021understanding}. Rainbow DQN \citep{hessel2018rainbow} has been shown to perform well on the Atari Learning Environment benchmark \citep{bellemare2013arcade} by combining a set of techniques (``Rainbow techniques'') that are empirically successful at improving sample efficiency, including Double Q-learning \citep{hasselt2010double,vanhasselt2015deep}, dueling networks \citep{wang2016dueling}, prioritized experience replay \citep{schaul2015prioritized}, distributional value estimates \citep{bellemare2017distributional}, and noisy network exploration \citep{noisyExplore}. 

\subsection{Stablizing Q-Learning}
Direct bootstrapping with a learned parametrized function approximator can cause instability and overestimation~\citep{hasselt2010double,vanhasselt2015deep,fujimoto2018addressing,song2018revisiting,Kim2019DeepMellowRT,kumar2019stablizing,kumar2020discor}. To alleviate these effect, Double Q-learning~\citep{hasselt2010double,vanhasselt2015deep} decorrelates the action optimization and value estimation, replacing the overestimation due to Jensen's inequality with a moderate underestimation.  Twin-Q \citep{fujimoto2018addressing} attempts to reduce overestimation more directly, by taking the minimum target value over two learner networks.  Soft-maximal value targets, often involving the mellow-max operator \citep{ziebart2010modeling,rubin2012trading,fox2016taming,asadi2017alternative,Kim2019DeepMellowRT}, can reduce both the bias~\citep{Fox2019Toward} and variance~\citep{Kim2019DeepMellowRT} of target value estimates.  A weighted TD error has also been proposed to handle uncertainty in the training signal caused by error propagation through the self-referential update structure \citep{kumar2020discor} and by model uncertainty due to model expressiveness and limited data \citep{lee2021sunrise}.

\subsection{Ensemble Learning in RL}
Ensemble learning, namely training a set of more than one learners for the same task, is often used in RL to guide exploration, and in some methods to improve target value estimates.
The statistics of the ensemble can be used to assess model uncertainty or to produce a lower-variance estimate, compared to a single estimator. These benefits have been utilized in many RL techniques, such as in measuring the error accumulation in a learned dynamics model in model-based RL~\citep{chua2018deep}, evaluating a temperature for softer maximization in uncertain states \citep{Fox2019Toward,Liang2021Temporal}, lowering the target estimate variance \citep{DBLP:journals/corr/AnschelBS16,an2021uncertainty}, biasing exploration toward novel states \citep{chen2017ucb, osband16a, lee2021sunrise}, down-weighting loss for uncertain target values~\citep{lee2021sunrise}, and alleviating estimation bias by pessimistically estimating the target as the minimum over ensemble predictions \citep{fujimoto2018addressing,lan2020maxmin}. Sufficient ensemble diversity is crucial in all of these methods~\citep{sheikh2022maximizing}.

While all ensemble methods are similarly motivated, approaches closely related to our own are
Averaged-DQN and Ensemble-DQN \citep{DBLP:journals/corr/AnschelBS16} and EBQL~\citep{peer2021ensemble}. Like MeanQ, these methods use an ensemble mean as the target estimate, with Ensemble-DQN and EBQL maintaining an explicit ensemble and Averaged-DQN reusing past snapshots of the network parameters. 
Both Ensemble-DQN and EBQL train all ensemble members with the same experience mini-batch and mean target values, unlike MeanQ which has each member sample independently from a shared replay buffer. The significance of these design choices is further discussed in \cref{avg_dqn_detail}.

Deep Exploration methods, such as RLSVI \citep{osband16b}, collect training data by selecting an ensemble member to interact with the environment throughout each episode. The ultimately deployed policy has the same property, unlike MeanQ which selects greedy actions using the ensemble mean in both exploration and deployment. Consistently with this difference, the methods also differ in how they generate target values, with RLSVI bootstrapping each member from its own value estimates and MeanQ averaging the ensemble for all targets. Thus, both methods collect data relevant to their targets and evaluated policies, which has been shown imperative for successful training~\citep{ostrovski2021difficulty}. We find empirically that this latter consideration outweighs the potential benefit of independent experience through Deep Exploration; that off-policy training with a replay buffer~\citep{mnih2015human} provides sufficient experience diversity in MeanQ; and that the correlation between ensemble members introduced by the shared replay buffer is alleviated by independent sampling from the buffer.

SUNRISE \citep{lee2021sunrise} and MeanQ differ only in how they compute the TD error. SUNRISE bootstraps each ensemble member from its corresponding target network, while MeanQ uses the ensemble mean. SUNRISE also rescales the TD error based on an ensemble-induced uncertainty measure of the target value.

\section{MeanQ}

We present MeanQ, a simple and sample-efficient ensemble-based RL algorithm. In this section, we discuss in detail each step of the algorithm and compare it with existing methods. Finally, we discuss how MeanQ can be combined with several RL techniques that have been shown useful in sample-efficient learning of Q networks.

\subsection{Updating Q Ensemble}

MeanQ estimates state–action values with an ensemble of $K$ Q networks. As in DQN, each network $k = 1, \ldots, K$, parametrized by $\theta_k$, takes a state $s$ as input, and outputs a vector predicting $Q_{\theta_k}(s, a)$ for each action $a$. The Q-network architecture allows maximizing the Q function over actions and finding a maximizing action in two operations that require it: (1) computing target values for optimization; (2) selecting a greedy action for rolling out the trained agent, whether to collect more data or to evaluate. In MeanQ, the maximum is taken over the mean value of the $K$ ensemble members:
$$V_\theta(s) = \max_a\{ \mean_k Q_{\theta_k}(s, a)\}.$$

In particular, this $V_\theta(s)$ is used for the target value when taking gradient steps to minimize the square TD error for each estimator $\theta_k$, with respect to target values computed by separate target networks $\bar{\theta}_1, \ldots, \bar{\theta}_K$
\begin{equation*}
    \mathcal{L}(s,a,r,s';\theta_k) = \bigg(r + \gamma V_{\bar{\theta}}(s') - Q_{\theta_k}(s,a) \bigg) ^ 2.
\end{equation*} 

\subsection{Decorrelating Ensemble Members}

The motivation for using an ensemble mean is that, when the members are not fully correlated, the mean has lower variance. We expect this variance reduction to provide the benefits discussed in \cref{section_reduce_variance}, including decreasing the overestimation bias in $V_{\bar{\theta}}(s')$ due to Jensen's inequality.

Effectively reducing variance requires keeping the ensemble members as uncorrelated as possible. However, complete independence of the members is impossible; in order to make use of the improved value estimates, they must affect each other's target values. MeanQ strikes a balance between keeping sources of estimate stochasticity independent when it can be done efficiently, and otherwise allowing their dependence. The complete MeanQ method is presented in Algorithm \ref{pseudo_code}.

The first source of estimate stochasticity is the initialization of each Q network~\citep{osband16a}, which is naturally done independently for each of the $K$ ensemble members.

When rolling out an exploration policy, it is possible for each $Q_{\theta_k}$ to form its own $\epsilon$-greedy exploration policy. However, that policy would not reflect our best available policy, resulting in suboptimal exploration. It would also differ from the evaluated policy
$$\pi_\theta(s) = \argmax_a\{ \mean_k Q_{\theta_k}(s, a)\},$$
which can be harmful~\citep{ostrovski2021difficulty}. We therefore use the ensemble mean for $\epsilon$-greedy exploration, by selecting $\pi_\theta(s)$ in state $s$ with probability $1 - \epsilon$ and a uniform action otherwise. 

Q-learning networks typically use a replay buffer to store experience and draw mini-batches for training.
This diversifies each mini-batch by mixing in it steps from different episodes or disparate times in an episode.
The next design choice for MeanQ is whether ensemble members use individual replay buffers (which could be experienced by $\pi_\theta$ or $\pi_{\theta_k} = \argmax_a Q_{\theta_k}$) or share a single buffer.
For the same amount of total exploration, a shared buffer is more diverse, and we find it empirically beneficial.

Using a shared experience policy and a shared replay buffer further correlates the ensemble members. To alleviate this negative effect and help the members evolve more independently, MeanQ has each member sample its mini-batches independently from the replay buffer. 
A downside, however, is computational efficiency.
Computing targets for each member requires us to evaluate all members' value estimates, $Q_{\theta_k}$, at each state $s'$ present in any of the members' mini-batches, performing 
$O(K^2)$ target Q function evaluations for $O(K)$ gradient updates.  
For small $K$ (5 in our experiments), this increase is not prohibitive, and is offset by the resulting improvement in sample efficiency, particularly in settings where environment interactions are more expensive than function evaluations.
\label{avg_dqn_detail}

MeanQ is closely related to Averaged-DQN (Avg-DQN) and Ensemble-DQN (Ens-DQN)~\citep{DBLP:journals/corr/AnschelBS16} and Ensemble Bootstrapped Q-Learning (EBQL)~\citep{peer2021ensemble}, which are similarly motivated.  
For computational reasons, these methods were designed to have members share mini-batch data (Ens-DQN and EBQL) or simply reuse recent snapshots of a single network (Avg-DQN). 
We find empirically that these choices degrade performance.
In our experiments (\cref{section_no_rainbow_exp}), we test Ens-DQN by using the same replay samples across members; intuitively, this makes the members more correlated and so reduces variance less.  Our experiments indicate that, while Ens-DQN provides only modest benefits over Avg-DQN, by modifying the ensemble training procedure to that of MeanQ we are able to improve sample efficiency significantly over Ens-DQN.

Finally, for prioritized experience replay~\citep{schaul2015prioritized}, the priority weights that control the sampling distribution are computed using the previous TD error for each data point. 
Independently sampling for each ensemble member enables us to also keep separate priorities, so that each member samples experience that is most relevant to its own value estimates.
To this end, MeanQ computes the priority weights for each ensemble member using its own previous TD errors.

\subsection{Combining with Existing Techniques}\label{sec:tricks} 
Our proposed method can be combined with several existing techniques for improving deep Q-learning, namely Rainbow DQN \citep{hessel2018rainbow} and UCB exploration \citep{chen2017ucb} as discussed below, and doing so further improves its performance.

\paragraph{Distributional target estimation.}  A useful modification to standard Q-learning is to have the network output a \emph{distribution} over values for each action \cite{bellemare2017distributional}. This value distribution is typically approximated by a categorical distribution over a fixed finite set of possible return values, called atoms.
A distributional MeanQ target estimate can be computed as an average over the members' estimates of the probability mass of each atom
\begin{align*}
    p_{s'} &= \mean_k p_{\bar{\theta}_k}(s',a^*) \\
    a^* &= \argmax_{a'}\{ \mean_k z^\intercal p_{\bar{\theta}_k}(s',a'),\}
\end{align*}
where $z$ are the locations of the return value's discrete support, similarly to \citet{bellemare2017distributional}.
Finally, we take member $k$'s loss function to be the cross-entropy loss between $p_{\theta_k}(s,a)$ and $p_{s'}$, with the latter projected onto the support $z$.  Implementation details are given 
in \cref{rainbow_meanq_algo}.

\paragraph{UCB exploration.} Ensembles have previously been used to assist in crafting an exploration policy \citep{auer2002finite,osband16a,chen2017ucb}. Since MeanQ already has an ensemble, it is easy to combine it with any of these exploration methods. In our experiments, we combine MeanQ with the exploration policy used in \citet{lee2021sunrise}
\begin{equation*}
    \pi(s)=\argmax_{a} \{ \mean_k Q_{\theta_k}(s,a) + \lambda \std_k Q_{\theta_k}(s,a) \},
\end{equation*}
where $\mean_k$ and $\std_k$ are the ensemble mean and standard deviation and $\lambda > 0$ is a hyperparameter.
\paragraph{Other techniques.} MeanQ can also trivially incorporate dueling network \citep{wang2016dueling} and noisy exploration~\citep{noisyExplore}, which have been observed to improve sample efficiency. 
MeanQ can also incorporate Double DQN \citep{hasselt2010double}, itself a size-2 ensemble method, but this would halve the effective ensemble size, and is therefore note done in our experiments.

\begin{algorithm}[t]
     \caption{MeanQ} \label{pseudo_code}
    Initialize $K$ Q networks $Q_{\theta_k}$ for all $k$ \\ 
    Initialize $K$ target Q networks $Q_{\bar{\theta}_k}$, $\bar{\theta}_k \leftarrow \theta_k$ for all $k$ \\
    Initialize replay memory $D$ to capacity $N$ \\
    Initialize prioritization \textcolor{blue}{$p_k(b|D)$} for all $k$ \\
     \For{$t=1, \ldots, T$}{
      Sample action $a_t$ according to exploration policy $\pi_\theta$ \\
      Observe $r_t \gets r(s_t,a_t)$ \\
      Sample $s_{t+1} \sim p(s_{t+1}|s_t,a_t)$ \\
      Store transition $(s_t, a_t, r_t, s_{t+1})$ in $D$ \\
      $s_{t} \leftarrow s_{t+1}$ \; \\
      \For{$k=1, \ldots, K$} 
          {Sample $B$ transitions $(s_b, a_b, r_b, s_b') \sim D$ \\
          \textcolor{blue}{$V_{s_b'}=\max_{a_b'}\{ \mean_k Q_{\bar{\theta}_k}(s_b',a_b') \}$} \\
          $y_{b} = \begin{cases}
          r_b & \text{if $s_{b}'$ is a terminal state} \\
          r_b + \gamma V_{s_b'} & \text{otherwise}
          \end{cases} $\\
          $\mathcal{L}(\theta_k) = \mean_b \mathcal{L}_b(\theta_k)$, \\ 
          where $\mathcal{L}_b(\theta_k) = (y_{b}-Q_{\theta_k}(s_b, a_b))^2$\\
          Update $\theta_k$ using $\nabla_{\theta_k} \mathcal{L}(\theta_k)$ \\
          Update \textcolor{blue}{$p_k(b|D)$} using \textcolor{blue}{$\sqrt{\mathcal{L}_b(\theta_k)}$}
        }
        Every $T_{\text{target}}$ steps: update ${\bar{\theta}}_{k} \leftarrow \theta_{k}$ for all $k$
     }
\end{algorithm}

\begin{figure*}[t]
    \includegraphics[width=\textwidth]{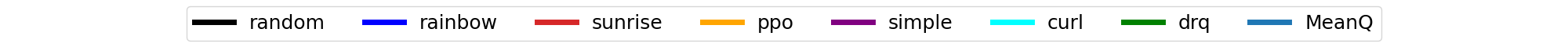}
    \includegraphics[width=0.195\textwidth]{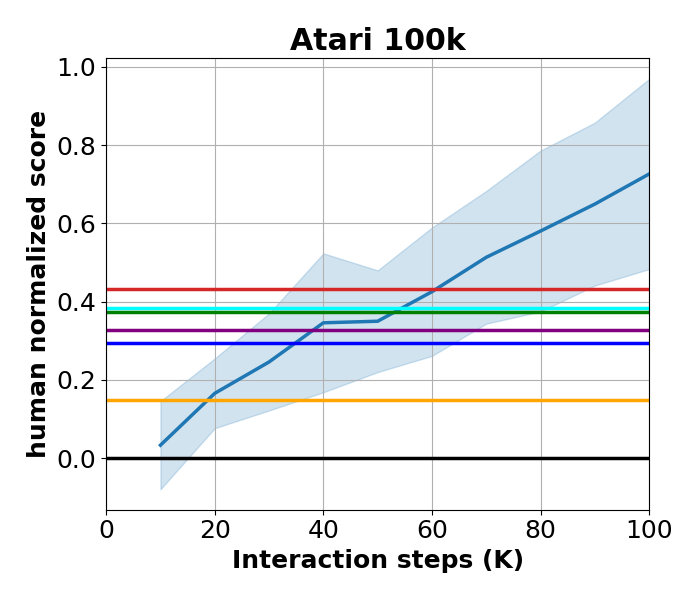}
    \includegraphics[width=0.195\textwidth]{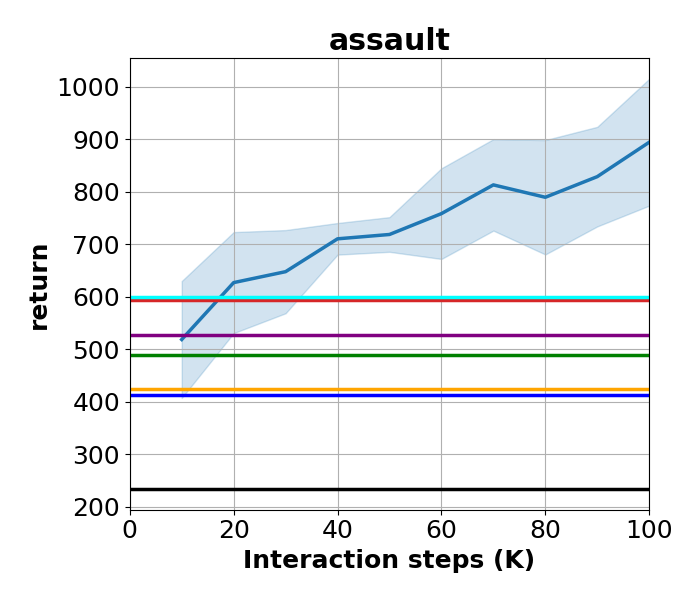}
    \includegraphics[width=0.195\textwidth]{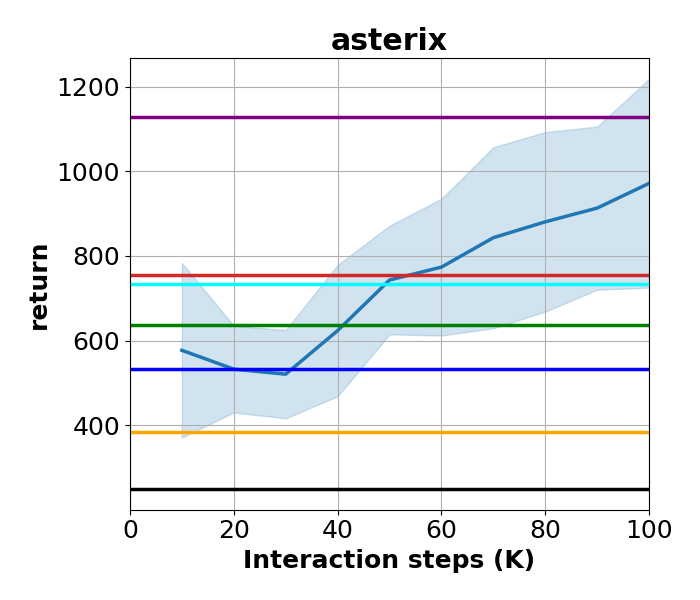}
    \includegraphics[width=0.195\textwidth]{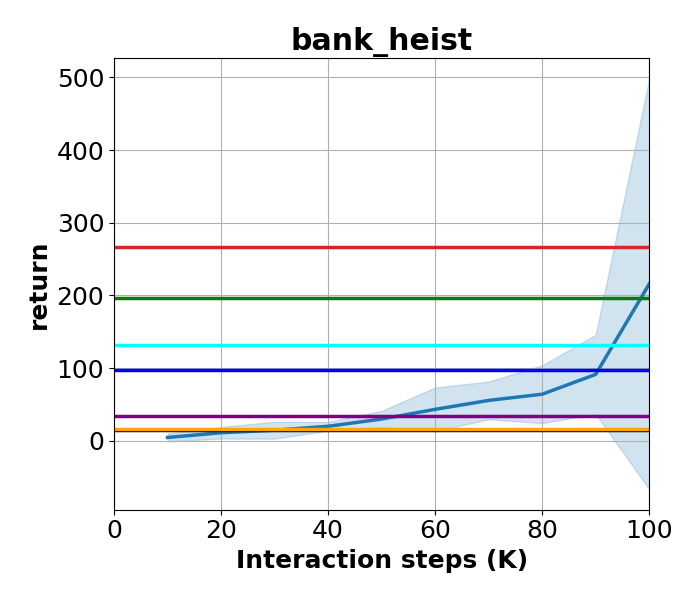}
    \includegraphics[width=0.195\textwidth]{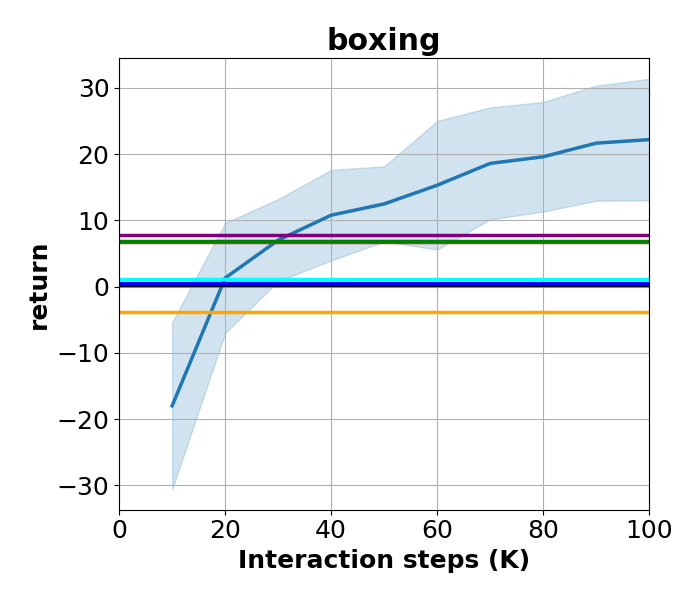}
    \includegraphics[width=0.195\textwidth]{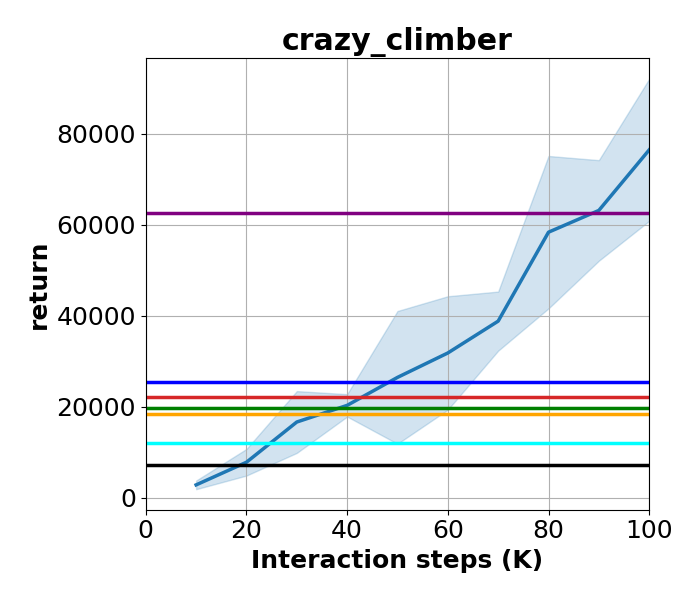}
    \includegraphics[width=0.195\textwidth]{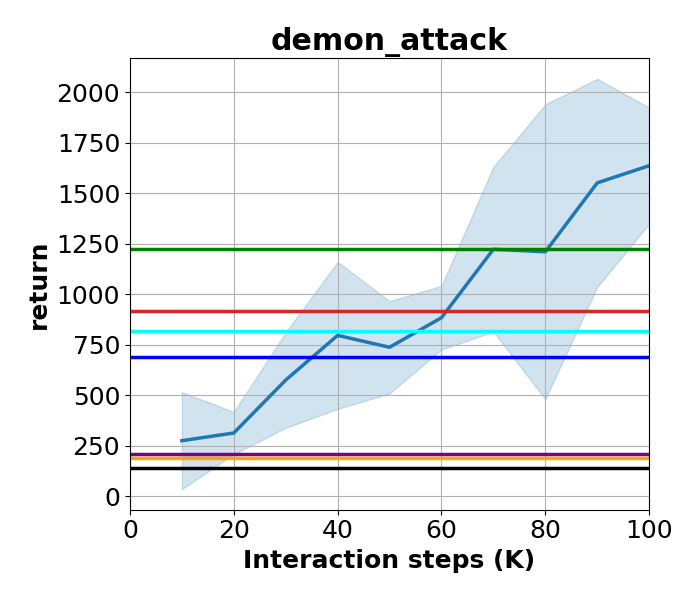}
    \includegraphics[width=0.195\textwidth]{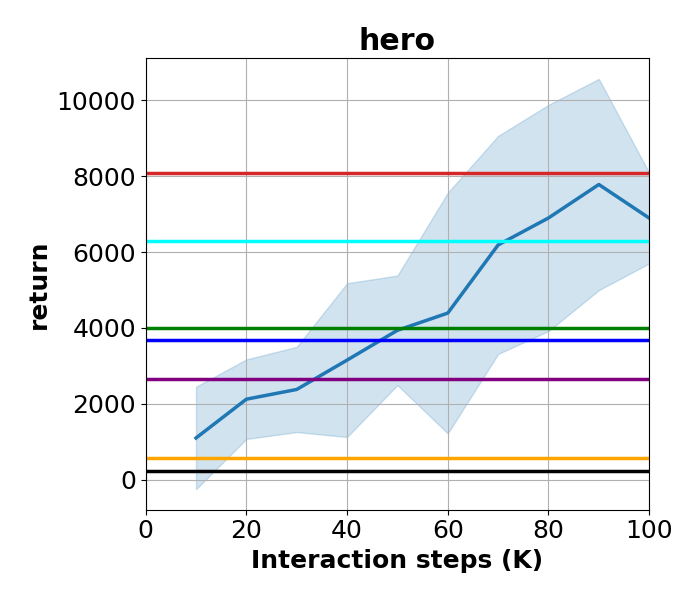}
    \includegraphics[width=0.195\textwidth]{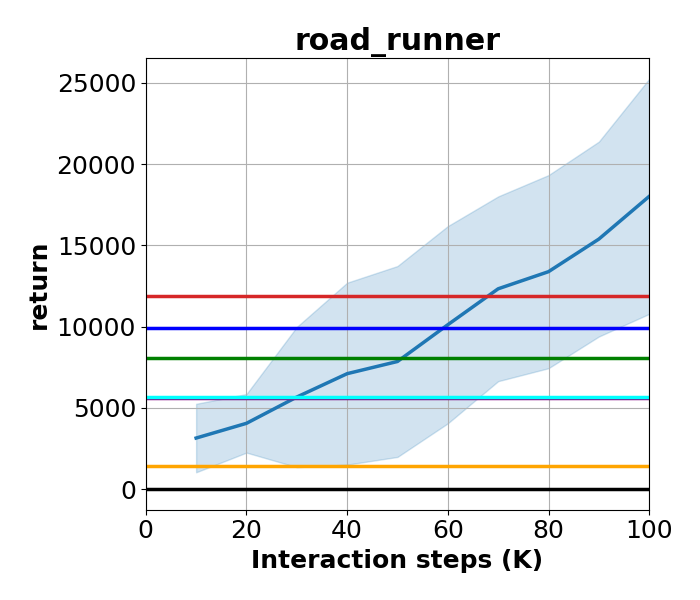}
    \includegraphics[width=0.195\textwidth]{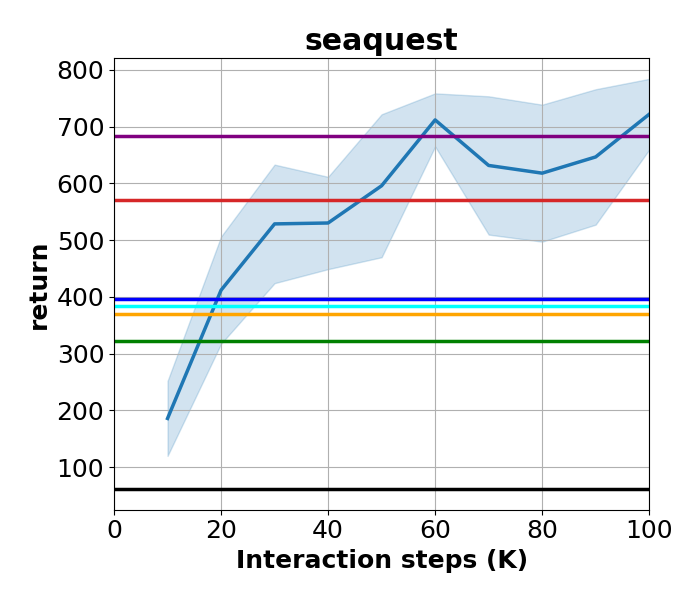}
    \caption{Performance of MeanQ with Rainbow techniques, compared with 6 existing algorithms. Mean expected undiscounted return over 5 runs is shown (shaded: 1 standard deviation), averaged over 26 Atari environments and in 9 individual ones. Baseline results of different algorithms at 100K interactions are rendered as horizontal lines and cited from: Rainbow \citep{van2019use}; SUNRISE~\citep{lee2021sunrise}; PPO, SimPLe, CURL, and DrQ \citep{kaiserModelBased}. Full results in all 26 environments are available in Appendix \ref{appendix_atari_results}. }
    \label{some_main_results_100k}
\end{figure*}

\section{Experiment Results} \label{all_exp}
 \begin{figure*}[t]
 \begin{center}
   \includegraphics[width=\textwidth]{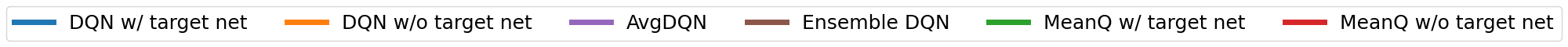}
   \includegraphics[width=0.195\textwidth]{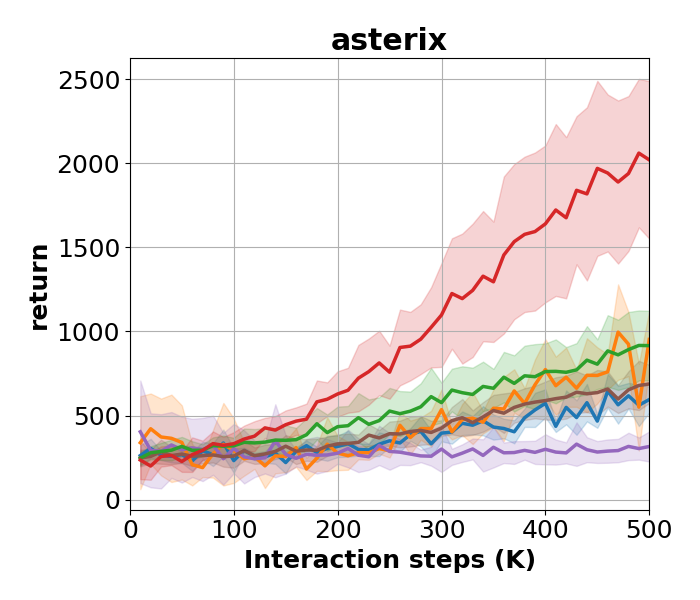}
   \includegraphics[width=0.195\textwidth]{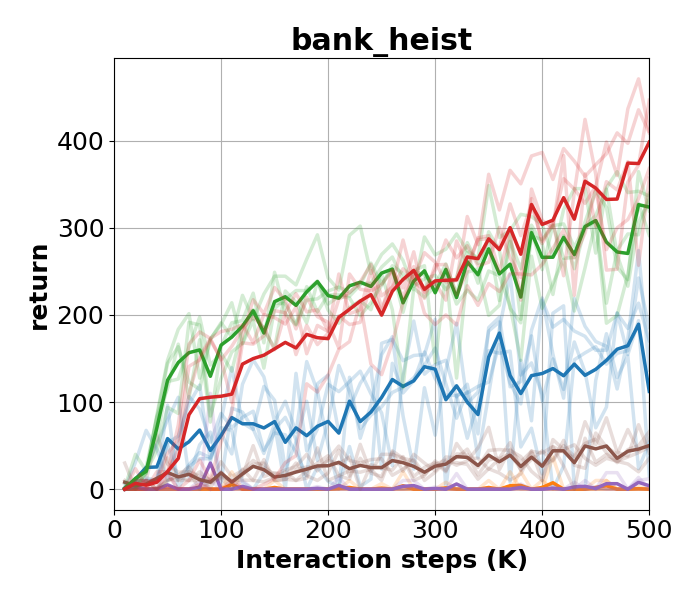}
   \includegraphics[width=0.195\textwidth]{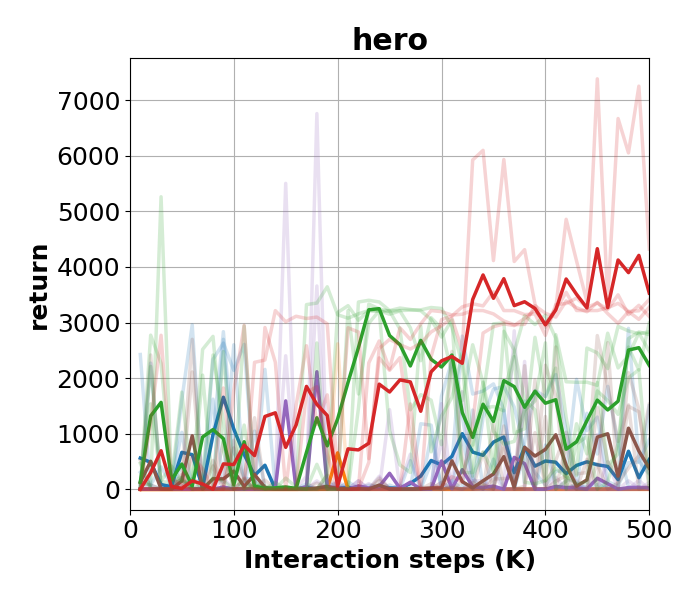}
   \includegraphics[width=0.195\textwidth]{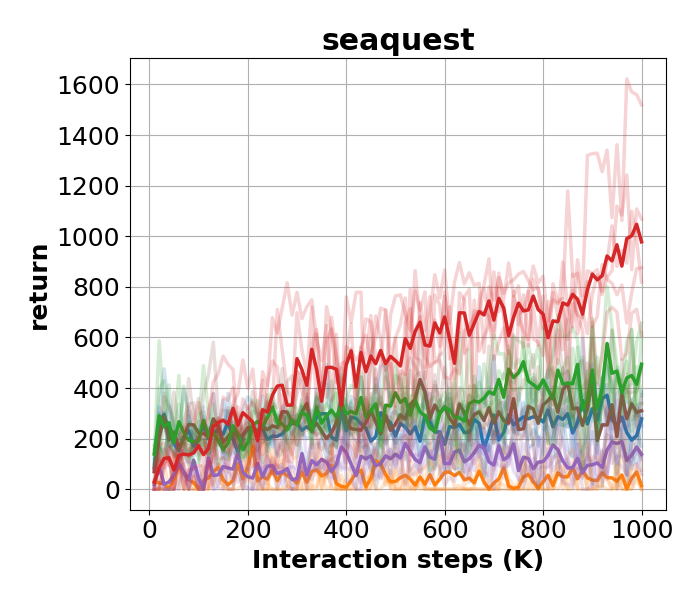}
   \includegraphics[width=0.195\textwidth]{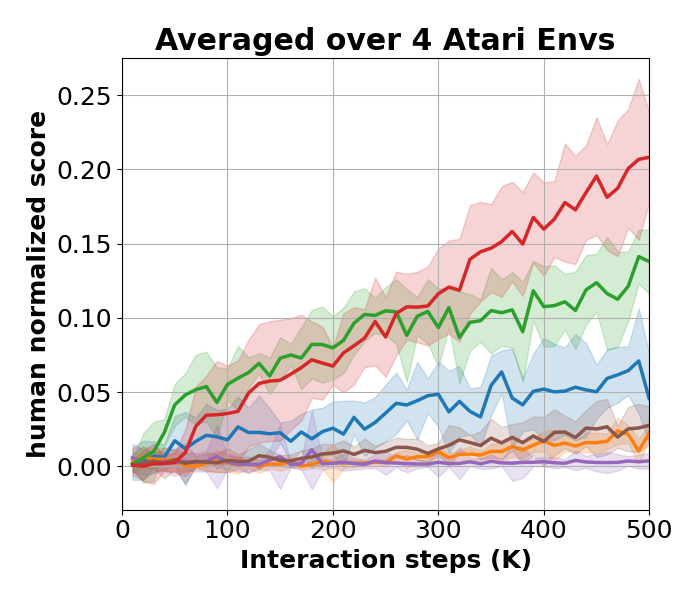}
   \includegraphics[width=0.195\textwidth]{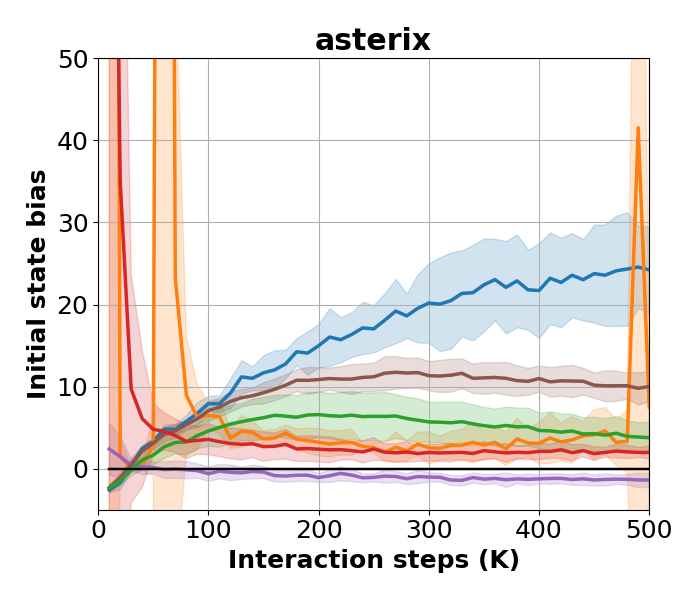}
   \includegraphics[width=0.195\textwidth]{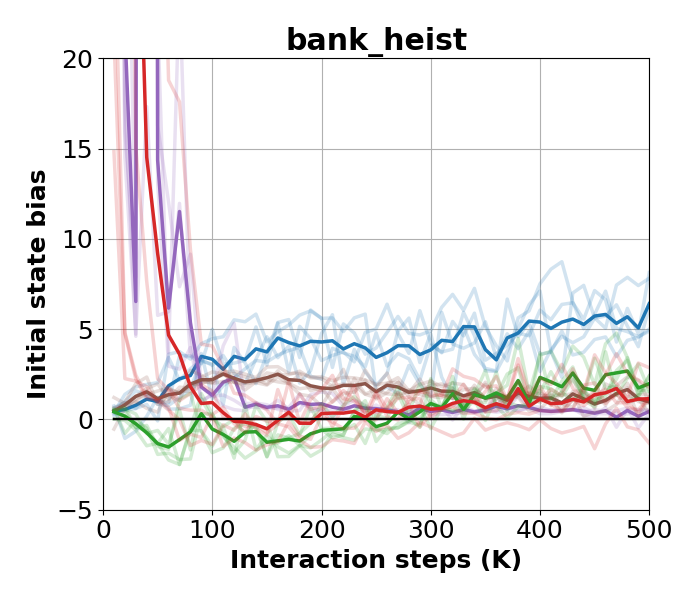}
   \includegraphics[width=0.195\textwidth]{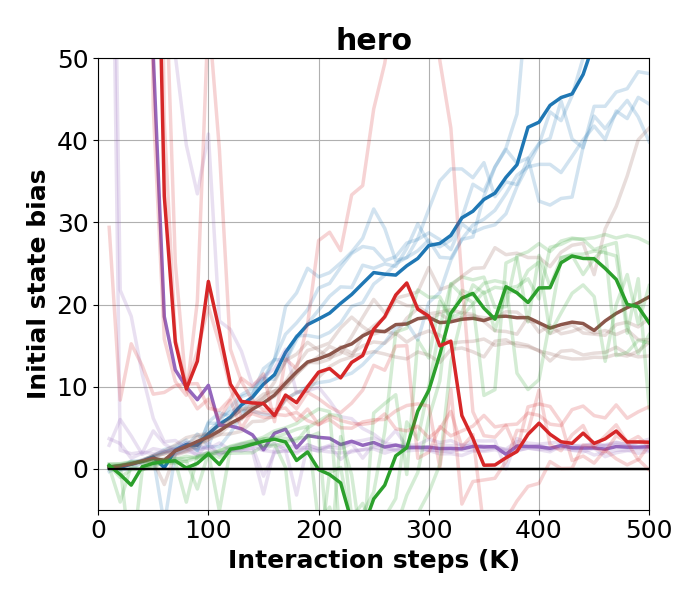}
   \includegraphics[width=0.195\textwidth]{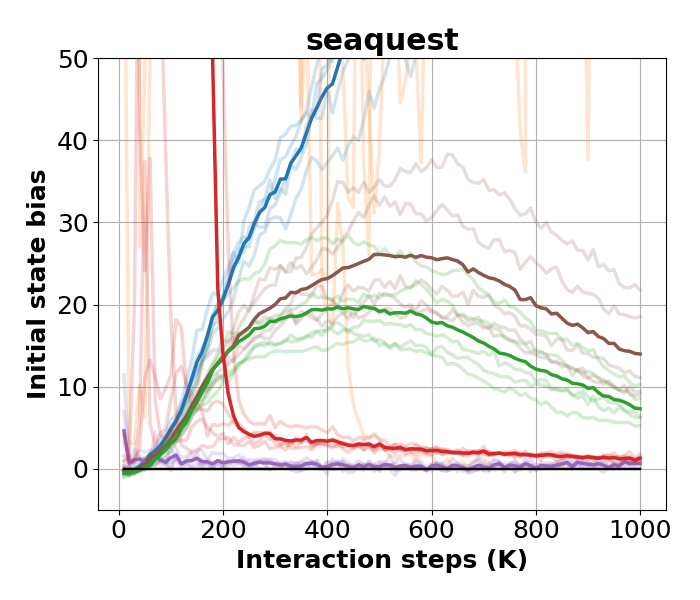}
   \includegraphics[width=0.195\textwidth]{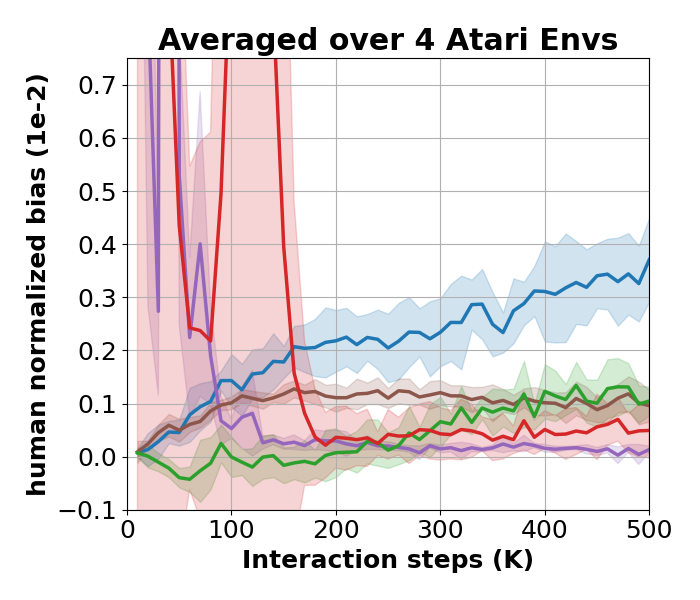}
      
   \includegraphics[width=0.196\textwidth]{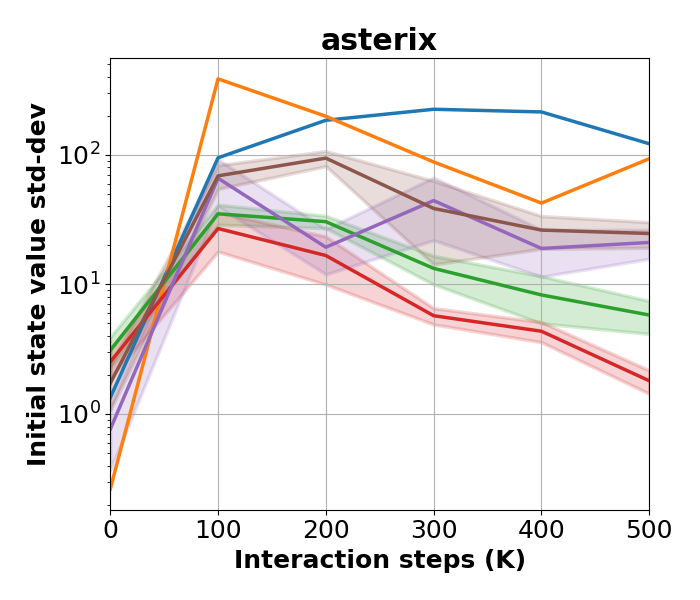}
   \includegraphics[width=0.196\textwidth]{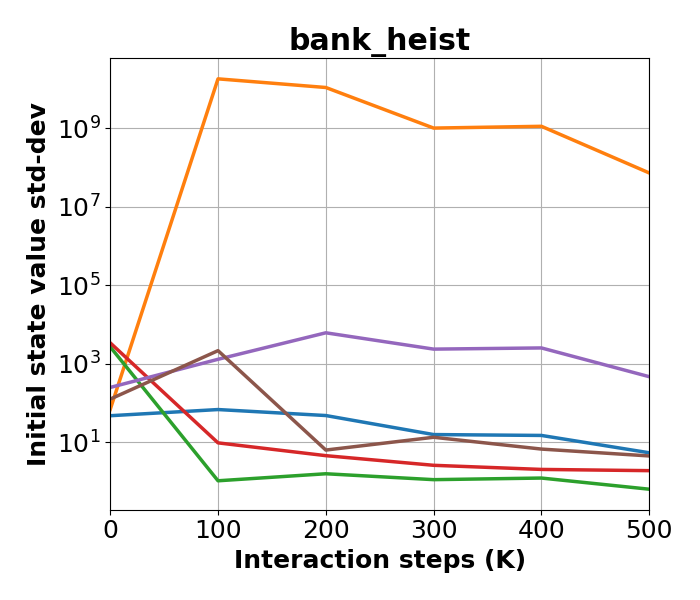}
   \includegraphics[width=0.196\textwidth]{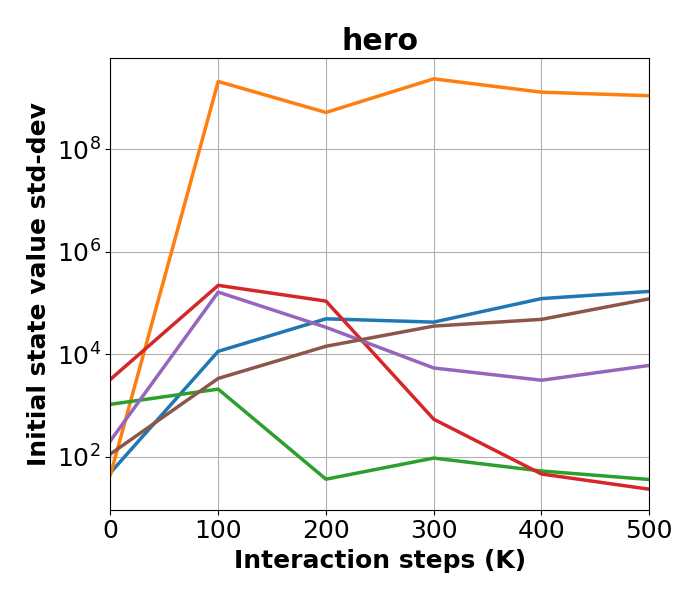}
   \includegraphics[width=0.196\textwidth]{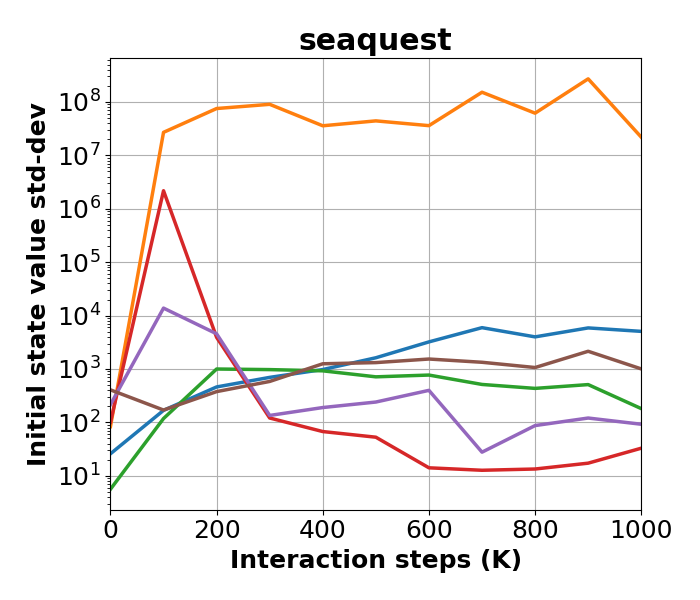}
   \includegraphics[width=0.196\textwidth]{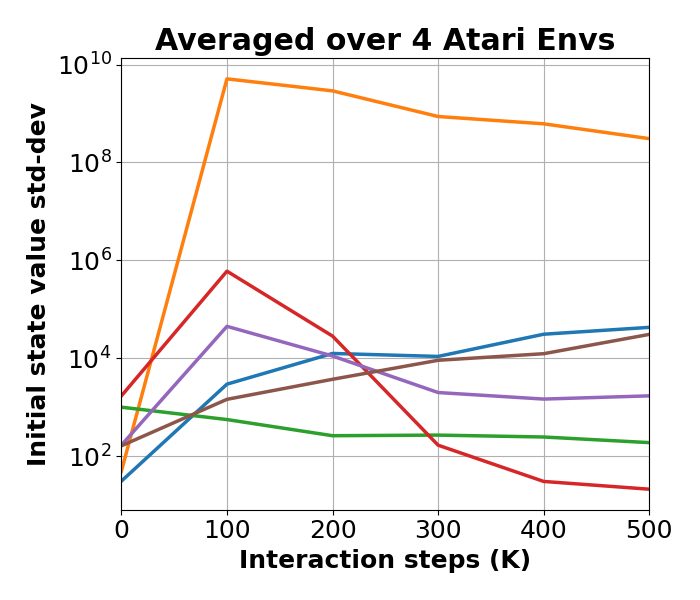}
   \includegraphics[width=0.196\textwidth]{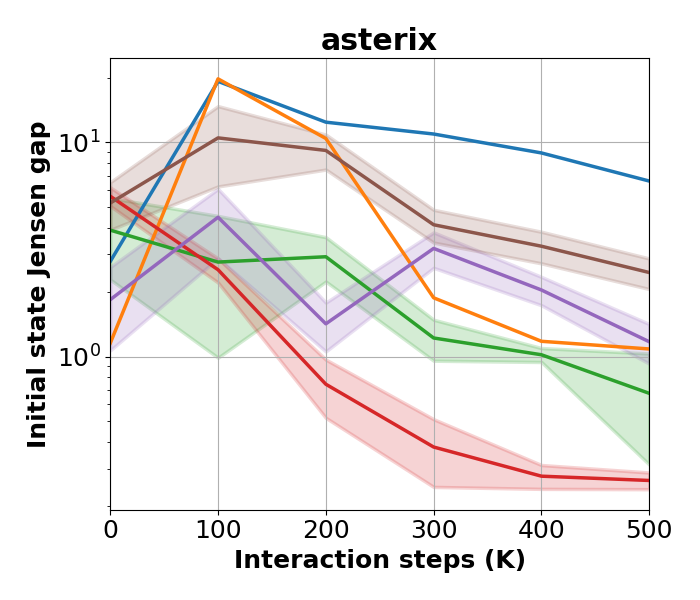}
   \includegraphics[width=0.196\textwidth]{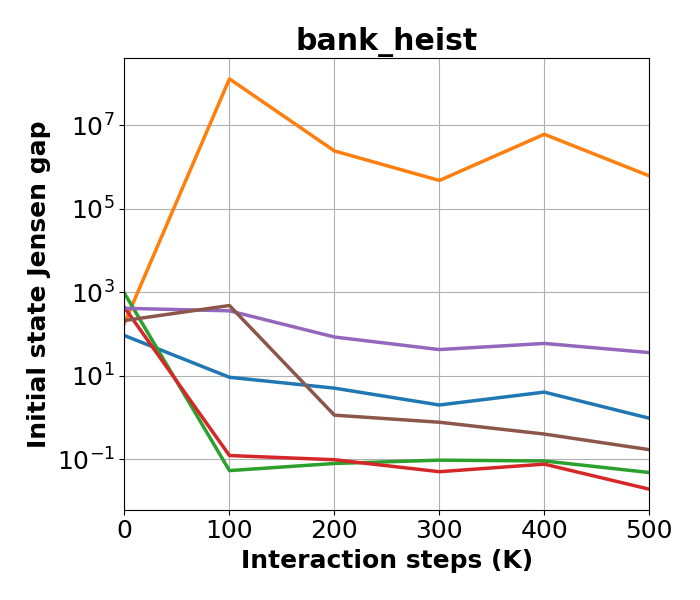}
   \includegraphics[width=0.196\textwidth]{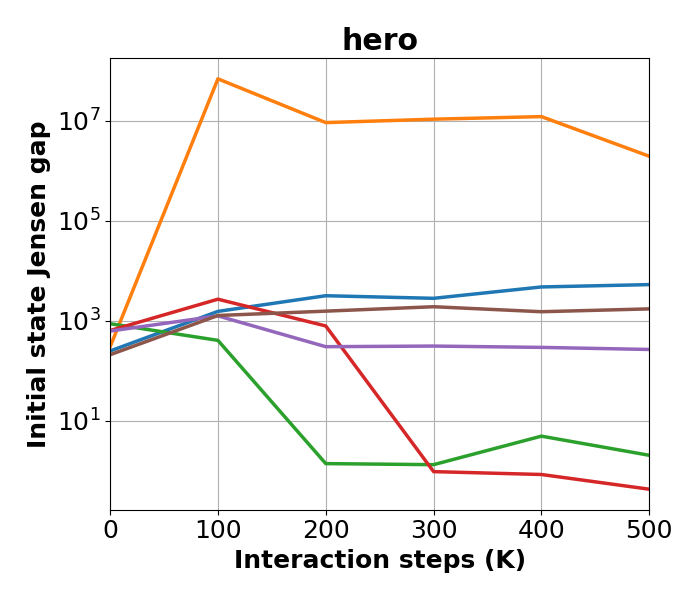}
   \includegraphics[width=0.196\textwidth]{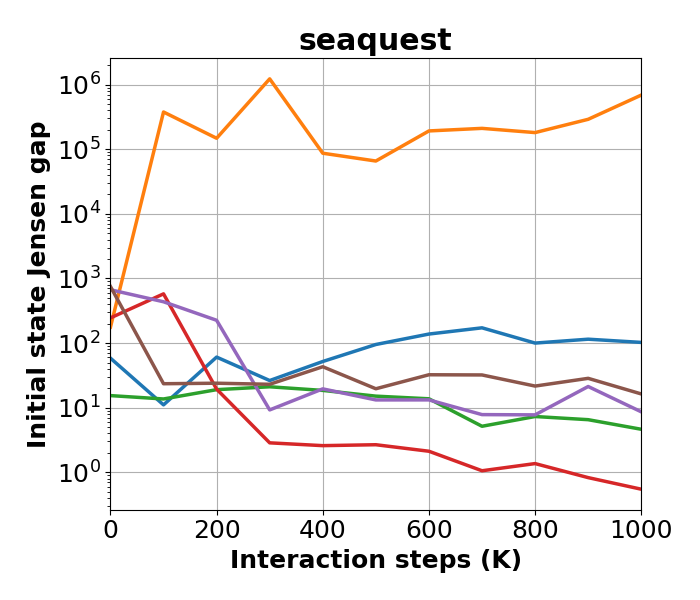}
   \includegraphics[width=0.196\textwidth]{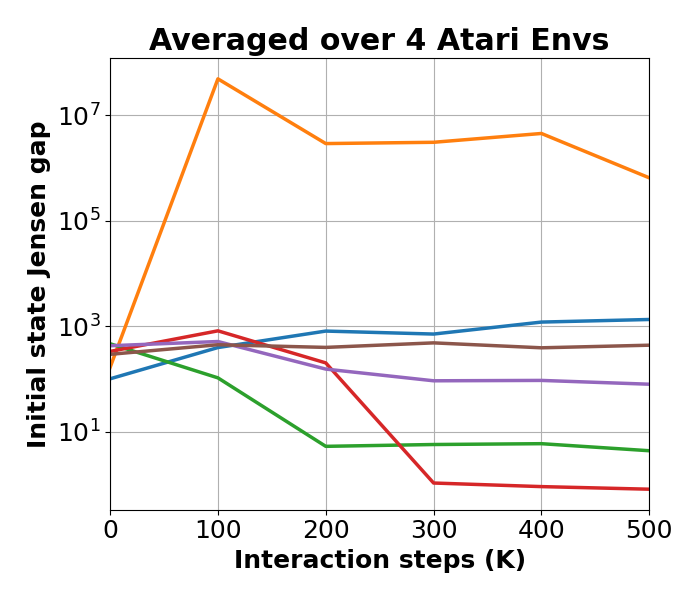}
  \caption{Performance, initial state estimation bias, standard deviation, and Jensen gap of DQN (with and without target network), Averaged-DQN, Ensemble-DQN, and MeanQ (with and without target network). Expected returns and biases, standard deviation, and Jensen gaps are over 5 runs; except in ensemble methods in the Asterix environment where 4 groups for 5 runs each are used to plot standard deviations (shaded; for details see text). Bias in DQN without target network is often above the plotted range. 
  In MeanQ, estimation variance is reduced by the ensemble averaging, eliminating the need for a target network. See \cref{section_no_rainbow_exp} for further analysis.} \label{no_rainbow_returns}
 \end{center}
\end{figure*}

\begin{figure*}[t]
    \begin{center}
    \includegraphics[width=\textwidth]{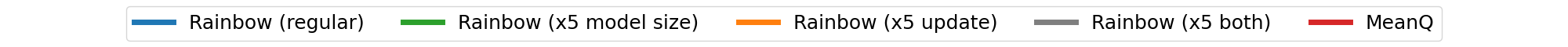}
    \includegraphics[width=0.245\textwidth]{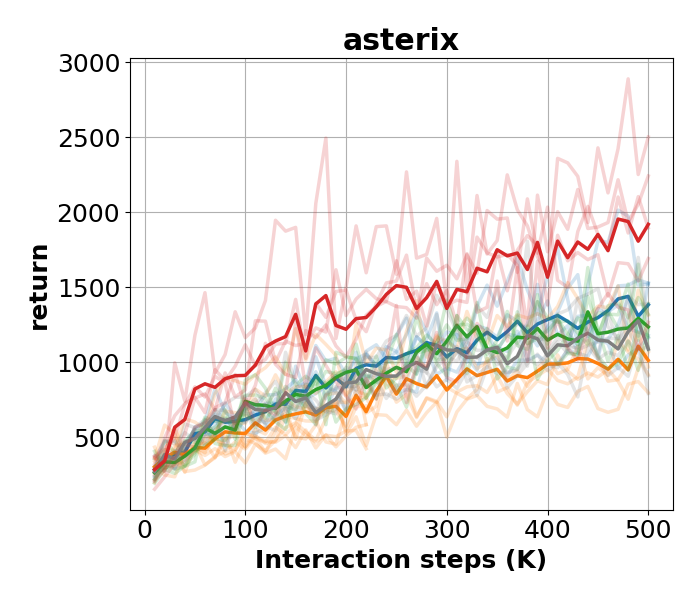}
    \includegraphics[width=0.245\textwidth]{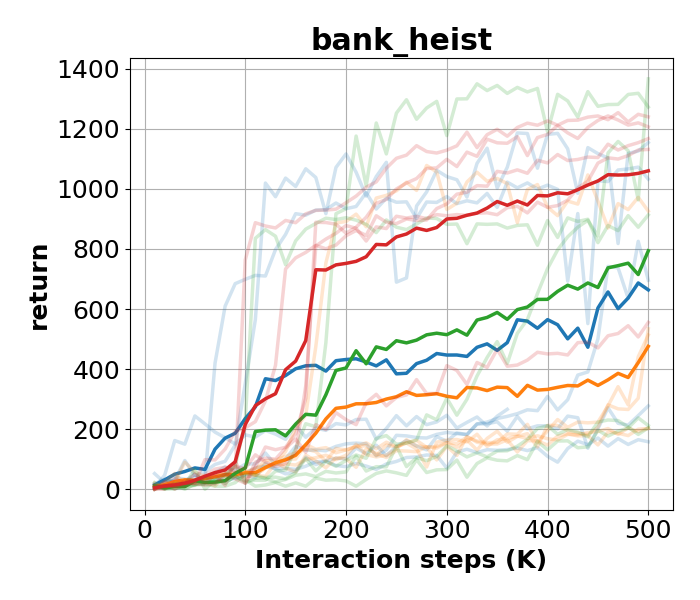}
    \includegraphics[width=0.245\textwidth]{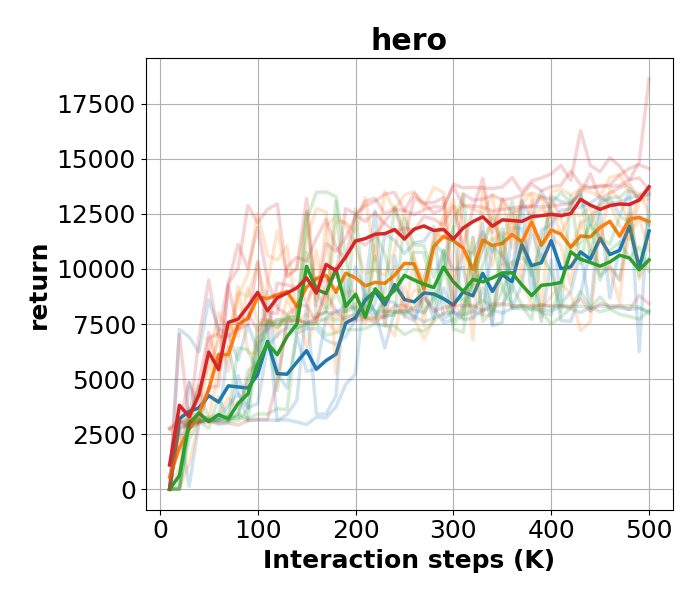}
    \includegraphics[width=0.245\textwidth]{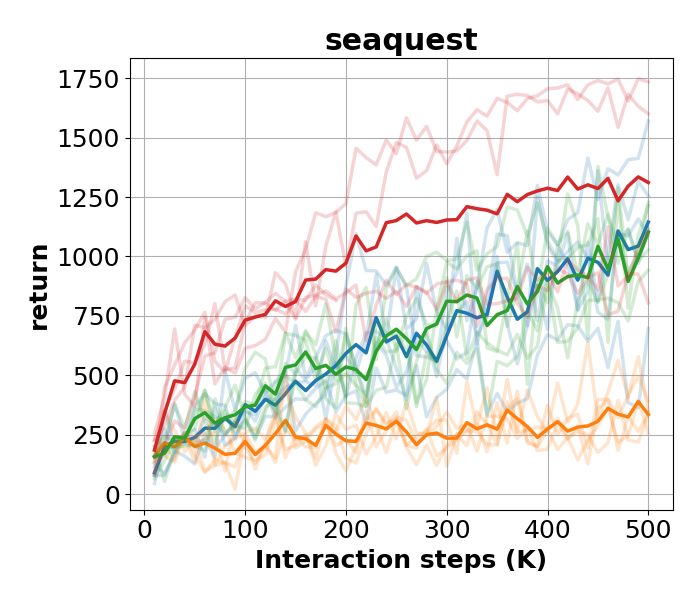}
    \includegraphics[width=0.245\textwidth]{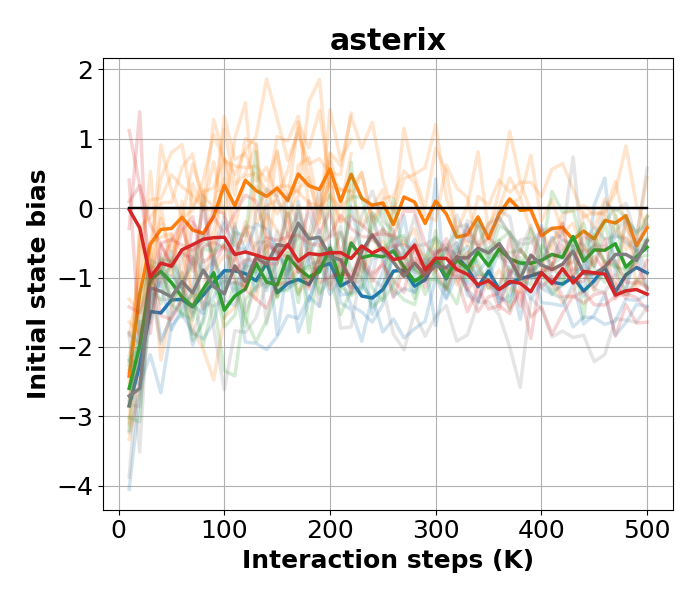}
    \includegraphics[width=0.245\textwidth]{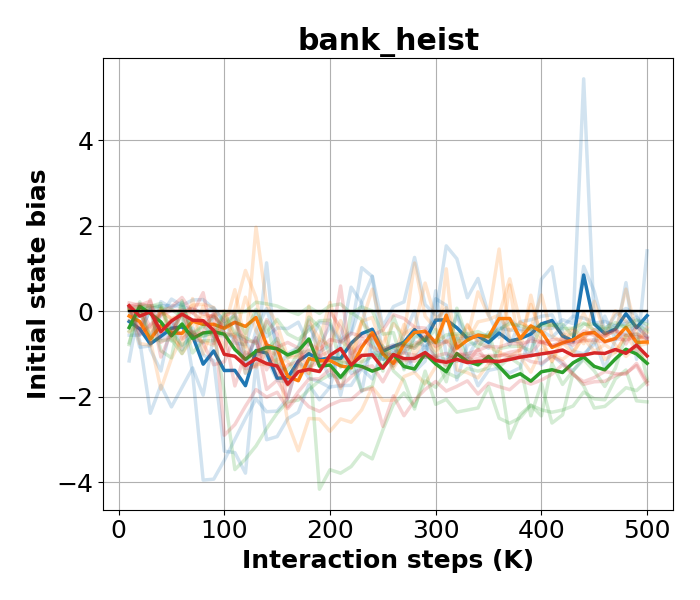}
    \includegraphics[width=0.245\textwidth]{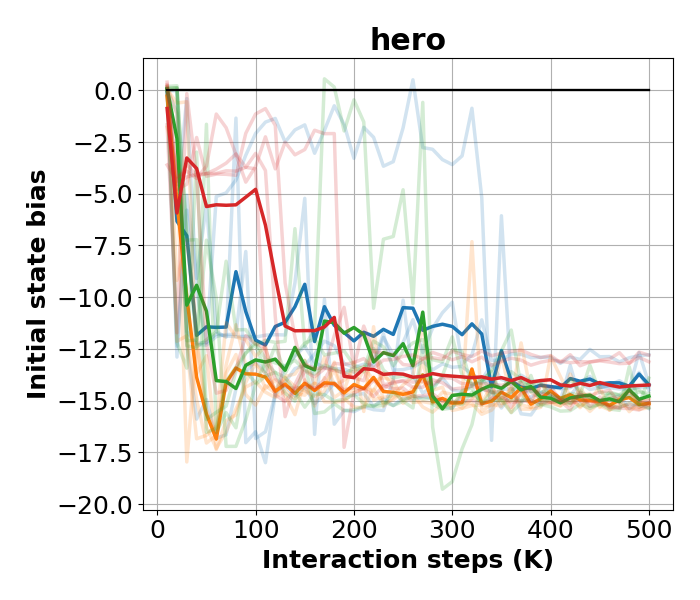}
    \includegraphics[width=0.245\textwidth]{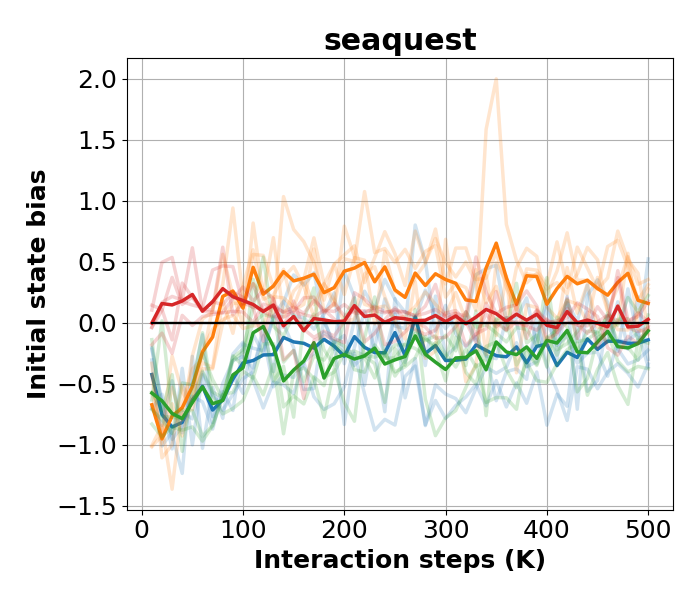}
    \caption{Performance of MeanQ with Rainbow techniques and UCB exploration, compared with several variations of Rainbow that resemble aspects of MeanQ: either a 5 times larger neural network, or 5 times more frequent network updates, or both in Asterix. }\label{rainbow_ablation}
    \end{center}
\end{figure*}

\subsection{Setup} 
We evaluate MeanQ on several discrete control benchmark environments in Atari \citep{bellemare2013arcade}. We compare with existing algorithms, including: (1) SUNRISE \citep{lee2021sunrise}, a model-free value-based ensemble RL method that is, to our knowledge, the state-of-the-art on this benchmark; 
(2) SimPLe \citep{kaiserModelBased}, a model-based method;
(3) PPO \citep{ppo}; (4) Rainbow DQN \citep{hessel2018rainbow} with two hyperparameter settings \citep{van2019use, kaiserModelBased}; (5) CURL \citep{laskin2020curl}; and (6) DrQ \citep{kostrikov2021image}. We also compare with human-level performance, as reported by \citet{kaiserModelBased}. For MeanQ, we use the version extended with the techniques of \cref{sec:tricks} and the same hyperparameter configuration as SUNRISE \citep{lee2021sunrise}, with two exceptions: no multi-step targets are used (compared with 3 steps in SUNRISE) and no target network is used. To remain comparable with SUNRISE, all our evaluations use an ensemble size $K=5$. Our MeanQ implementation is described in full detail in Appendix \ref{rainbow_meanq_algo} and available at \url{https://github.com/indylab/MeanQ}.

\subsection{Evaluation} \label{main_exp}
Our results in Atari environments are summarized in \cref{some_main_results_100k}. The full results in all 26 environments are available in Appendix \ref{appendix_atari_results}.
The experiment results show that MeanQ significantly outperforms all baselines with which we compared. MeanQ outperforms the best available baseline, SUNRISE, at 100K interaction steps in 16/26 environments. Scaling and shifting the returns such that random-policy performance is 0 and human-level performance is 1, to allow averaging across environments, MeanQ outperforms SUNRISE by 68\% on average over 5 runs in each of the 26 environments. MeanQ also achieves higher returns than Rainbow DQN at 500K steps in 21/26 environments, and 49\% higher average normalized return. Averaged over 5 runs in each environment, MeanQ achieves average human-level performance in the 26 games using only 200K ($\pm$100K) interaction steps.
We note that, since our method is compatible with many more existing techniques which we did not evaluate, such as contrastive learning \citep{laskin2020curl}, augmentation \citep{kostrikov2021image}, and various model-based methods, it has the potential for further performance improvement.

\subsection{Effect of Target Network}\label{section_no_rainbow_exp}

In this experiment, we compare DQN and MeanQ with and without a lagging target network. To isolate the effects on MeanQ itself, we experiment without including any of the techniques described in \cref{sec:tricks} (``vanilla MeanQ").  To study the effect of variance on overestimation bias and policy performance, it is also important not to use a distributional return representation~\citep{bellemare2017distributional}, which curtails overestimation due to the clipping effect of projecting the value distribution onto a fixed discrete support. Exploration is $\epsilon$-greedy with linear annealing of $\epsilon$. The detailed modification in hyperparameter configuration for this experiment is available in Appendix~\ref{hyperparameters}.

To study the algorithmic behavior, we measure a set of statistics during training:
\begin{enumerate} \setlength{\itemsep}{-1pt}
    \item \textbf{Performance.} We measure performance by the undiscounted sum of rewards $\sum_t r_t$, averaged over 20 evaluation trajectories after every 10K interaction steps of the exploration policy. 
    \item \textbf{Estimation bias.} We measure the value estimation bias in the initial state $s_0$. In this experiment, a prediction of the learner network $V(s_0)=\max_a Q_\theta(s_0,a)$ is queried from the learner. For ensemble methods, $V(s_0)=\max_a\{ \mean_k Q_{\theta_k}(s_0,a)\}$ is queried. We then report the empirical mean of $V(s_0) - \sum_t \gamma^t r_t$ over 20 evaluation trajectories after every 10K interactions. 
    \item \textbf{Estimation variance.} We measure the variance of value estimates by reporting the standard deviation of the initial state value $V(s_0)$ over 5 runs of each algorithm after every 100K interactions.
    Scaling and shifting algorithm performance such that random-policy performance is 0 and human-level is 1, the standard deviation is normalized by this standardized performance to obtain the relative standard deviation.\footnote{This normalization method is only approximate, since value estimates are discounted and performance is undiscounted.} The relative standard deviation is averaged over 50 resets of $s_0$.
    \item \textbf{Jensen gap.} We measure the Jensen gap $\E[\max_{a} Q(s_0,a)] - \max_{a} \E[Q(s_0,a)]$, with the expectation estimated empirically in
    5 runs of each algorithm after every 100K interactions. The Jensen gap is normalized by standardized performance, and averaged over 50 resets of $s_0$.
\end{enumerate}

The quantities above are evaluated in initial states $s_0$, because these are states that all algorithms must learn to evaluate in every run. Other states may not have the same reach probabilities across algorithms and runs, and comparing them is less indicative of the actual algorithmic behavior. 

Figure \ref{no_rainbow_returns} summarizes the empirical estimates of performance, bias, variance, and Jensen gap of 6 algorithms: DQN \citep{mnih2015human}, DQN without a target network \citep{DBLP:journals/corr/MnihKSGAWR13}, Avg-DQN and Ens-DQN \citep{DBLP:journals/corr/AnschelBS16}, MeanQ, and MeanQ without a target network.
Rows 1 and 2 show the mean performance and bias over 5 runs, as well as all runs (faded); except in ensemble methods in the Asterix environment, where the mean and 1 standard deviation (shaded) over 20 runs are shown; and the average over 4 environments in the last column, where the mean and 1 standard deviation (shaded) are shown.
Rows 3 and 4 show the standard deviation and Jensen gap over 5 runs; except in ensemble methods in Asterix, where the 20 runs are split into 4 groups and the mean and 1 standard deviation (shaded) of the within-group statistics are shown.
The results suggest the following findings:
\begin{enumerate} \setlength{\itemsep}{-1pt}
    \item Across all methods and variations, higher estimation variance and higher Jensen gap are positively correlated.
    \item Reproducing known results, a target network \citep{mnih2015human} in DQN significantly reduces overestimation bias (often too high to plot) and estimation variance \citep{hasselt2018deep}.
    \item MeanQ outperforms DQN, with and without a target network. MeanQ also has lower estimation variance, Jensen gap, and overestimation bias.
    \item As in DQN, removing the target network in MeanQ tends to cause a mild increase in estimation variance, overestimation bias, and Jensen gap, and decrease in performance in early-stage training. However, in a reversal of the findings in DQN, MeanQ without a target network quickly overtakes MeanQ with a target network in terms of normalized variance and Jensen gap, which translates into a significant bias reduction and performance gain.
    \item While Avg-DQN and, to a lesser extent, Ens-DQN reduce the variance and bias compared with DQN, this does not generally correspond to performance gain. We note that it is easier to keep bias low when performance is low, because bias is relative to returns and because values are easier to estimate when near-constant. 
    \item MeanQ with and without a target network does better than Ens-DQN in all four considered metrics (with the exception of bias with a target network). This demonstrates the importance of independent replay sampling and prioritization for learning neural network ensembles.

\end{enumerate}

\subsection{Ensembles vs. Larger Model and More Updates}
\label{ablation_exp}

In this experiment, we rule out two alternative hypotheses for the source of performance improvement in MeanQ. One could wonder whether the improvement could come from the ensemble being an effectively larger model or from performing more frequent value updates per interaction step. In this section, we empirically answer these questions in the negative. 

\paragraph{Larger model.} We experiment with Rainbow DQN with the same hyperparameters and a neural network architecture equivalent to the ensemble used in MeanQ. This network consists of 5 parallel networks, each taking input $s$ and outputting $Q_{\theta_k}(s,a)$. The final layer is a fixed mean over $Q_{\theta_k}(s,a)$ with no additional parameters. The results in \cref{rainbow_ablation} show no significant improvement compared to Rainbow DQN with a standard network. 

\paragraph{More frequent updates.} We experiment with Rainbow DQN with the same hyperparameters, except with 5 times as many gradient steps per interaction step. The results in~\cref{rainbow_ablation} show, in all environments but in some more than others, a drop in performance, compared with the baseline hyperparameters, caused by performing gradient steps as frequently in Rainbow DQN as in MeanQ with 5 ensemble members. An experiment in the Asterix environment (Figure \ref{rainbow_ablation}, left) also shows no benefit from combining both ablations.

These experiments reaffirm our motivation of reducing value estimate variance through ensemble averaging, and the significant role this plays in improving sample efficiency. MeanQ's improvements cannot be attributed to the larger model or more gradient updates.

\section{Conclusion}
In this paper, we introduce MeanQ, a simple ensemble RL method that uses the ensemble mean for target value estimates. Three key design choices, which set MeanQ apart from the extensive prior study of ensemble RL, are: (1) a shared exploration policy, likewise computed from the ensemble mean; (2) a shared replay buffer for all ensemble members; and (3) independent replay sampling from the buffer. While (1) and (2) can further correlate the ensemble members, beyond their use for each other's updates, we find that (1) is necessary to collect relevant experience, (2) helps experience diversity, and (3) is sufficient to maintain ensemble diversity. Averaging a sufficiently diverse ensemble reduces the variance of target value estimates, which in turn mitigates overestimation. Variance reduction in MeanQ also stabilizes training enough to obviate the lagging target network, which eliminates another source of estimation bias. Our experiment results show improved performance over record-holding baseline algorithms, as well as ablative versions.

Our empirical study furthers our understanding of the role of variance in value-based reinforcement learning, but much remains unknown about the interplay between variance, bias, exploration, and the deadly triad of off-policy bootstrapping with function approximation. Of particular interest is the necessity of nonlinear function approximation for the benefit of MeanQ to manifest, as learning in linear and tabular representations cannot benefit from a simple mean. We expect the community's further investigation to benefit from the simple and effective method proposed here, which can lend itself to more theoretical and empirical analyses, and serve as a stronger baseline for future work.

\section*{Acknowledgements}

The authors would like to thank Sameer Singh and Kimin Lee for their help.

The research work of authors Dailin Hu and, partly, Roy Fox is funded by the Hasso Plattner Foundation.

\bibliography{main}
\bibliographystyle{icml2022}

\newpage
\appendix
\onecolumn


\section{MeanQ with Rainbow Techniques and UCB Exploration} \label{rainbow_meanq_algo}
Results in Figures \ref{some_main_results_100k}, \ref{atari_100k}, \ref{atari_500k}, and Tables \ref{main_results_100k}, \ref{main_results_500k} are produced by the implementation of Algorithm \ref{rainbow_meanq_algo_pseudocode} in \url{https://github.com/indylab/MeanQ}.
\begin{algorithm}[h] 
    \caption{MeanQ with Rainbow techniques and UCB exploration} \label{rainbow_meanq_algo_pseudocode}
    Initialize return distribution support $z$ with $L$ atoms spaced evenly from $V_\text{min}$ to $V_\text{max}$ by $\Delta z$ \\
    Initialize $K$ parametrized return distribution $p_{\theta_k}$ for all $k$ \\
    Initialize replay memory $D$ to capacity $N$ \\
    Initialize prioritization \textcolor{blue}{$p_k(i|D)$} for all $k$ \\
     \For{$t=1, \ldots, T$} {
      Sample noisy network weights \\
      \textcolor{blue}{$a_t \gets \argmax_{a} \{ \mean_k Q_{\theta_k}(s,a)  + \lambda\std_k Q_{\theta_k}(s,a)  \}$} \\
      Observe $r_t \gets r(s_t,a_t)$ \\
      Sample $s_{t+1} \sim p(s_{t+1}|s_t,a_t)$ \\
      Store transition $(s_t, a_t, r_t)$ sequentially in $D$ \\
      $s_{t} \leftarrow s_{t+1}$ \\
        \For{$k=1, \ldots, K$}{
            Sample $B$ multi-step transitions $(s_b, a_b, r_b^{(0)}, ..., r_b^{(M - 1)}, s_b^{(M)}) \sim D$ \\
            \textcolor{blue}{$a^*=\argmax_{a} \{\mean_k z^\intercal p_{\theta_k}(s_b^{(M)},a)\}$} \\
            \textcolor{blue}{$p_{s_b^{(M)}}=\mean_k p_{\theta_k}(s_b^{(M)},a^*)$} (as a distribution vector over atoms) \\
            $c \gets \vec{0}$ \\
            $r_\text{M-step} = \sum_{m=0}^{M-1} \gamma^m r_b^{(m)}$ \\
            \For{$j=1, \ldots, L$}{
                $\mathcal{\hat{T}} z_{j}= 
                \begin{cases}
                  r_\text{M-step} & \text{if $s_{b}^{(M)}$ is a terminal state} \\
                   r_\text{M-step} + \gamma^M z_j & \text{otherwise}
                \end{cases} $\\
                $h_j = (\text{clip}[\mathcal{\hat{T}} z_{j}]_{V_\text{min}}^{V_\text{max}} - V_\text{min}) / \Delta z$ \\ 
                $l = \lfloor{h_j}\rfloor$, $u = \lceil{h_j}\rceil$ \\
                $c_l \gets c_l + p_{s_b^{(M)}}(u - h_j)$; $c_u \gets c_u + p_{s_b^{(M)}}(h_j - l)$
            }
            $\mathcal{L}(\theta_k) = \mean_b\mathcal{L}_b(\theta_k)$, where $\mathcal{L}_b(\theta_k) = - c^\intercal \log p_{\theta_k}(s_b, a_b) $ \\ 
            Update $\theta_k$ using $\nabla_{\theta_k}\mathcal{L}(\theta_k)$ \\
            Update \textcolor{blue}{$p_k(b|D)$} using \textcolor{blue}{$\mathcal{L}_b(\theta_k)$}
        } 
    }
\end{algorithm}

\section{Hyperparameters} \label{hyperparameters}
The implementation of MeanQ is adapted from a publicly released implementation repository (\url{https://github.com/Kaixhin/Rainbow}). For results in \cref{main_exp,ablation_exp}, our hyperparameters are the same as in \citet{lee2021sunrise}, except that no multi-step targets are used ($M=1$ instead of 3 in \citet{lee2021sunrise}). In \cref{main_exp}, a target network is not used for MeanQ. MeanQ does not introduce additional hyperparameters. For results in \cref{section_no_rainbow_exp}, we list the modified hyperparameters compared with \cref{main_exp} in \cref{hyper}. The algorithms with "w/ target net" variant in \cref{section_no_rainbow_exp} are using the same update frequency as in \cref{main_exp}.
\begin{table}[h]
\caption{Hyperparameter of Vanilla MeanQ in \cref{section_no_rainbow_exp}}\label{hyper}
\label{sample-table}
\vskip 0.15in
\begin{center}
\begin{tabular}{cc}
\toprule
Hyperparameter & Value \\
\midrule
Network architecture & same as \citet{wang2016dueling} \\
Learning rate $\alpha$    & $6.25 \times10 ^{-5}$  \\
Epsilon greedy $\epsilon$    & Linear annealing from $1$ to $0.1$ from $t=0$ to 200K  \\
Interactions per gradient update & Seaquest: 3 ; Asterix, BankHeist, and Hero: 2 \\
Multi-step ($M$) & 1 \\
\bottomrule
\end{tabular}
\end{center}
\vskip -0.1in
\end{table}

\section{Performance of 26 Atari Environments} \label{appendix_atari_results}
\begin{figure*}
    \includegraphics[width=\textwidth]{figures/main_100k/legend.png}
    \includegraphics[width=0.195\textwidth]{figures/human_normalized/human_100.png}
    \includegraphics[width=0.195\textwidth]{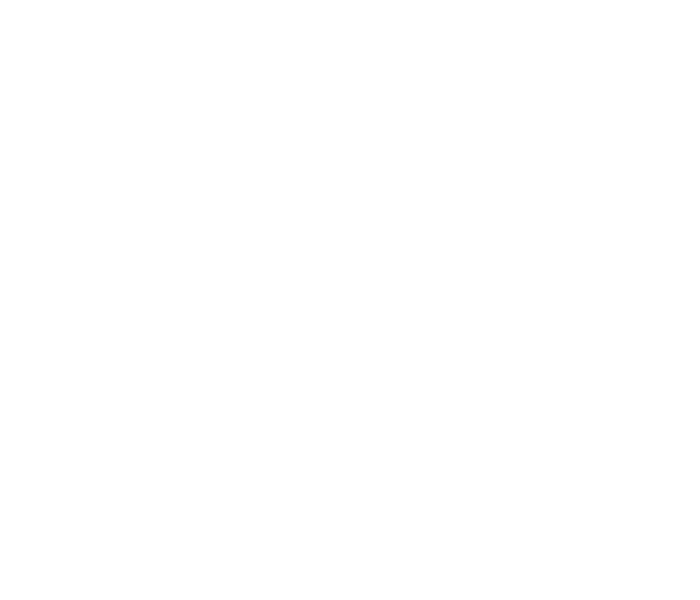}
    \includegraphics[width=0.195\textwidth]{figures/websiteplanet-dummy-700X600}
    \includegraphics[width=0.195\textwidth]{figures/websiteplanet-dummy-700X600}
    \includegraphics[width=0.195\textwidth]{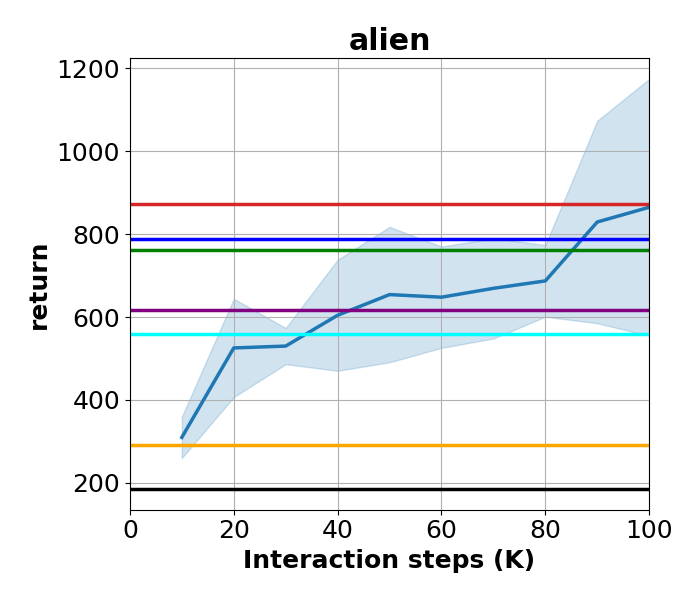}
    \includegraphics[width=0.195\textwidth]{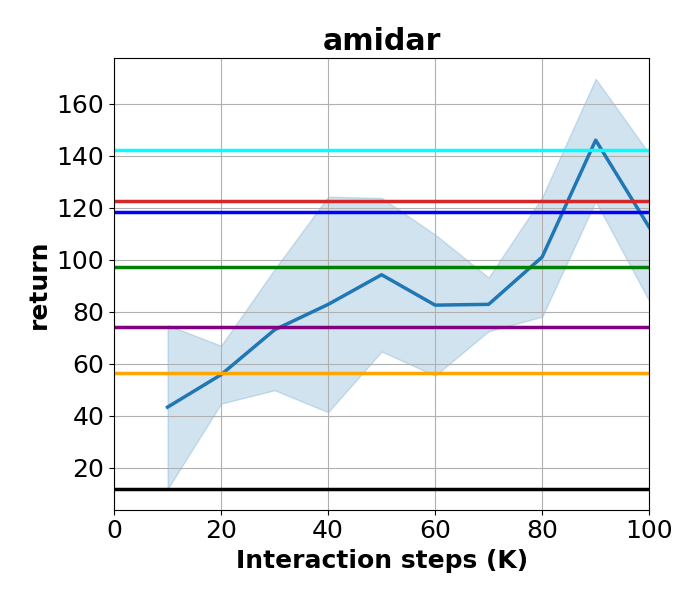}
    \includegraphics[width=0.195\textwidth]{figures/main_100k/assault_return.png}
    \includegraphics[width=0.195\textwidth]{figures/main_100k/asterix_return.png}
    \includegraphics[width=0.195\textwidth]{figures/main_100k/bank_heist_return.png}
    \includegraphics[width=0.195\textwidth]{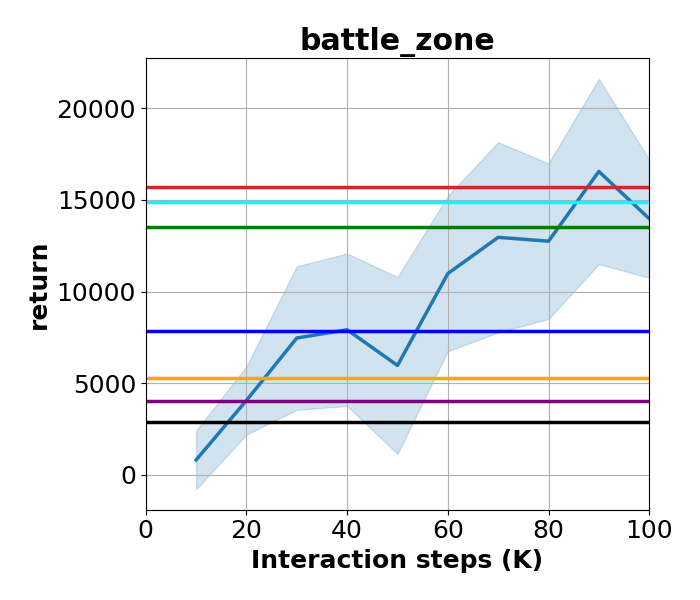}
    \includegraphics[width=0.195\textwidth]{figures/main_100k/boxing_return.png}
    \includegraphics[width=0.195\textwidth]{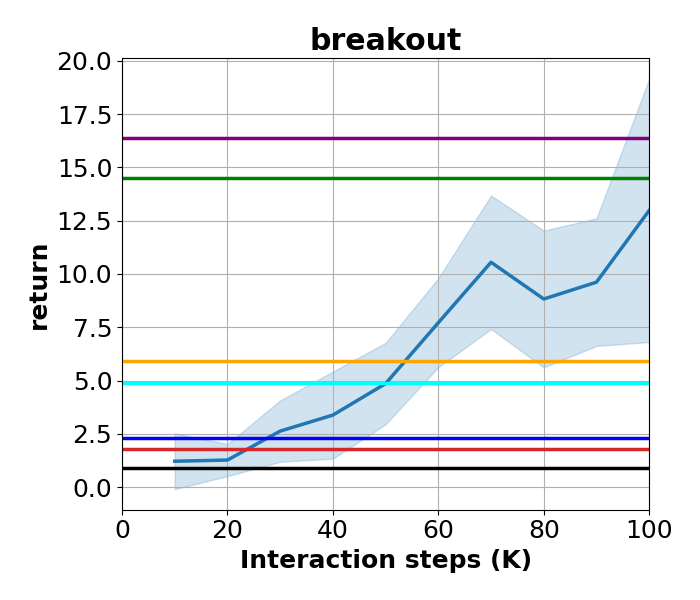}
    \includegraphics[width=0.195\textwidth]{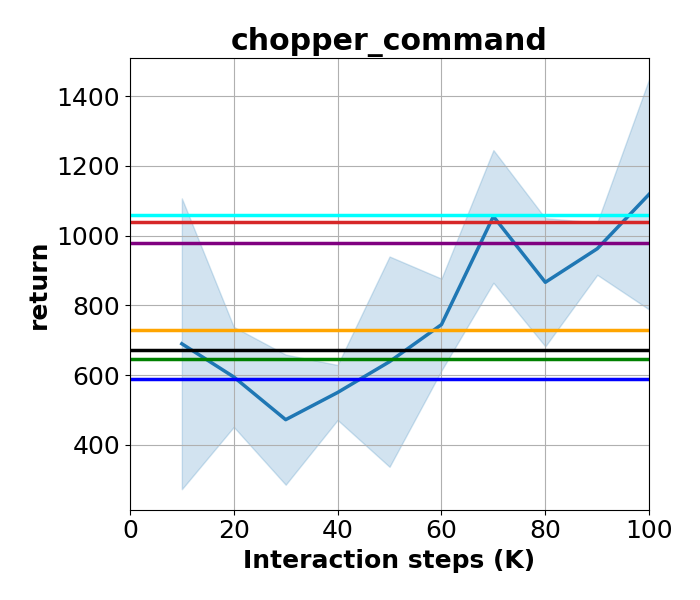}
    \includegraphics[width=0.195\textwidth]{figures/main_100k/crazy_climber_return.png}
    \includegraphics[width=0.195\textwidth]{figures/main_100k/demon_attack_return.png}
    \includegraphics[width=0.195\textwidth]{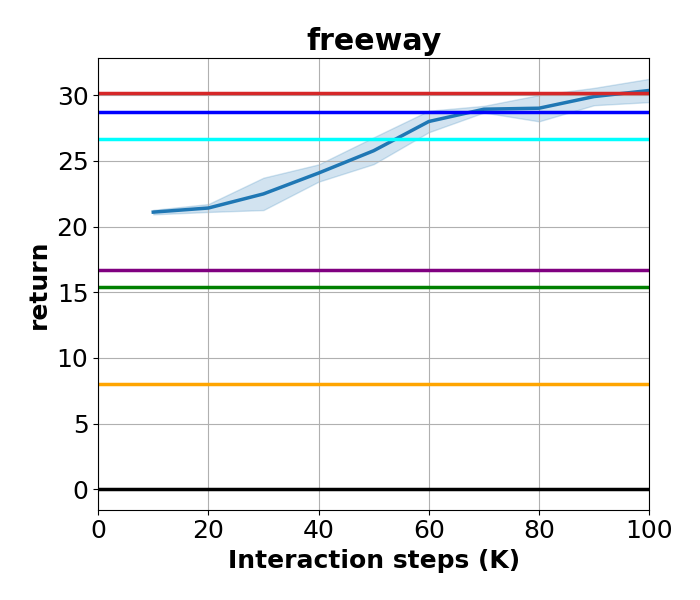}
    \includegraphics[width=0.195\textwidth]{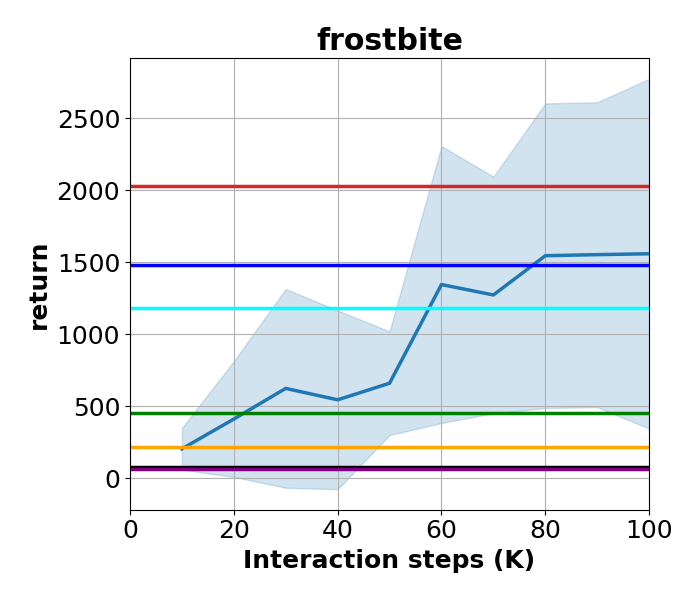}
    \includegraphics[width=0.195\textwidth]{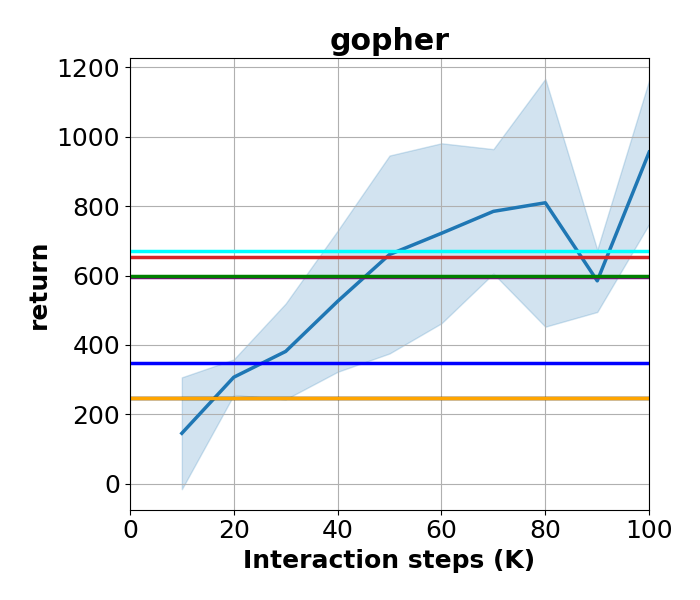}
    \includegraphics[width=0.195\textwidth]{figures/main_100k/hero_return.png}
    \includegraphics[width=0.195\textwidth]{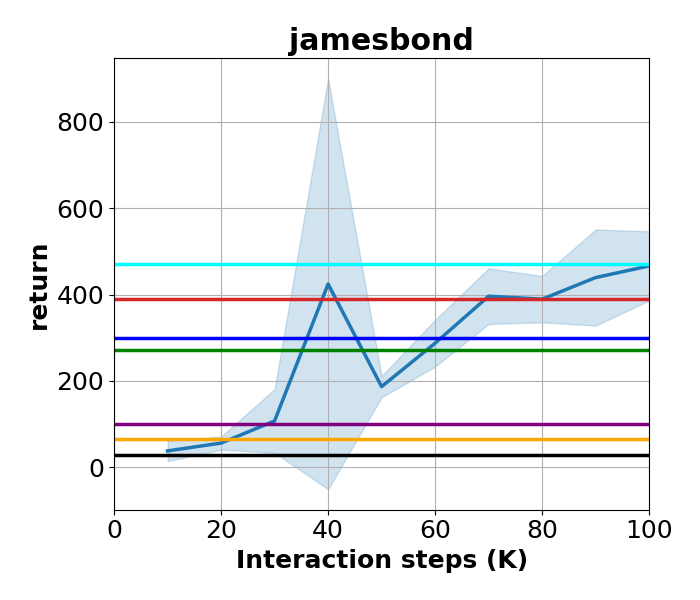}
    \includegraphics[width=0.195\textwidth]{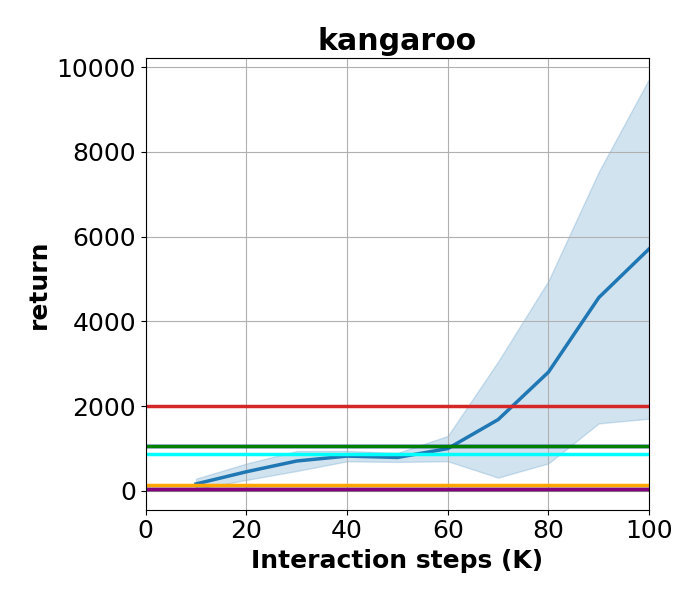}
    \includegraphics[width=0.195\textwidth]{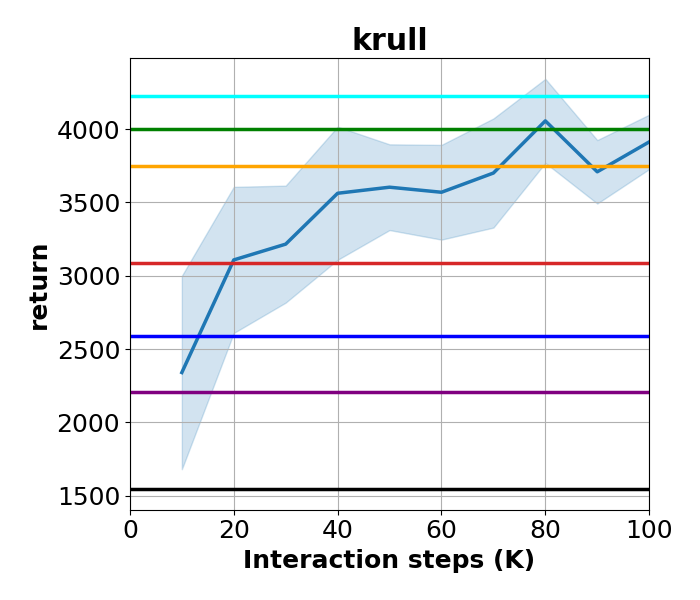}
    \includegraphics[width=0.195\textwidth]{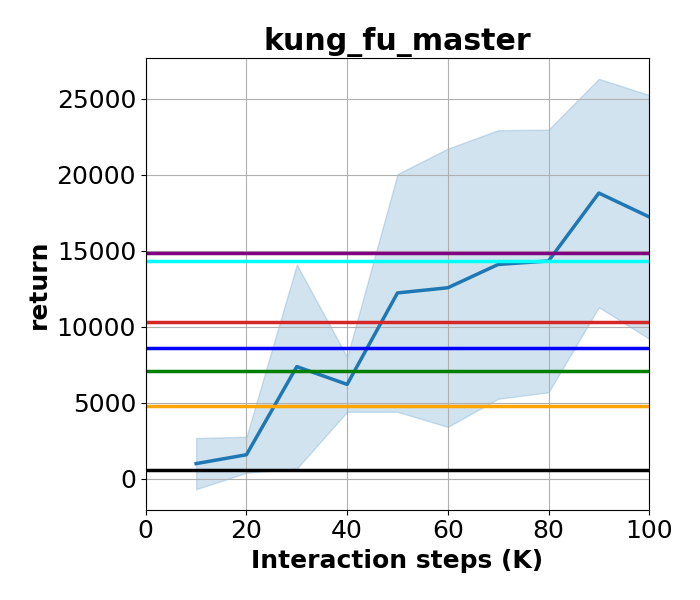}
    \includegraphics[width=0.195\textwidth]{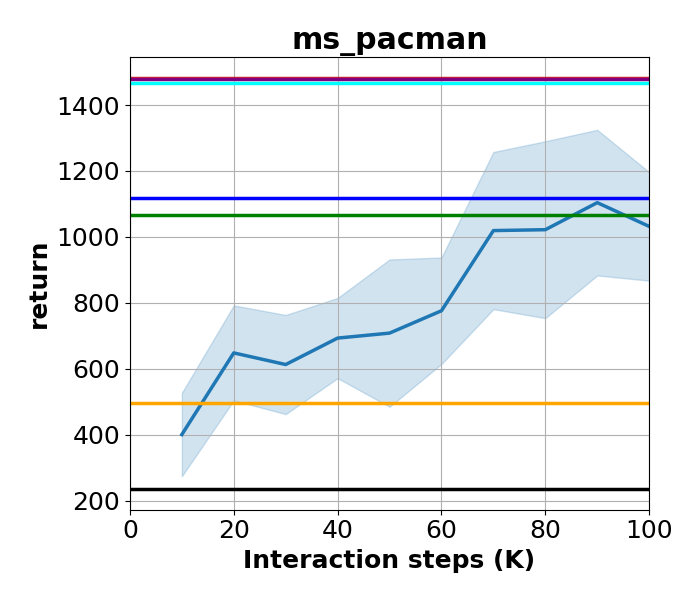}
    \includegraphics[width=0.195\textwidth]{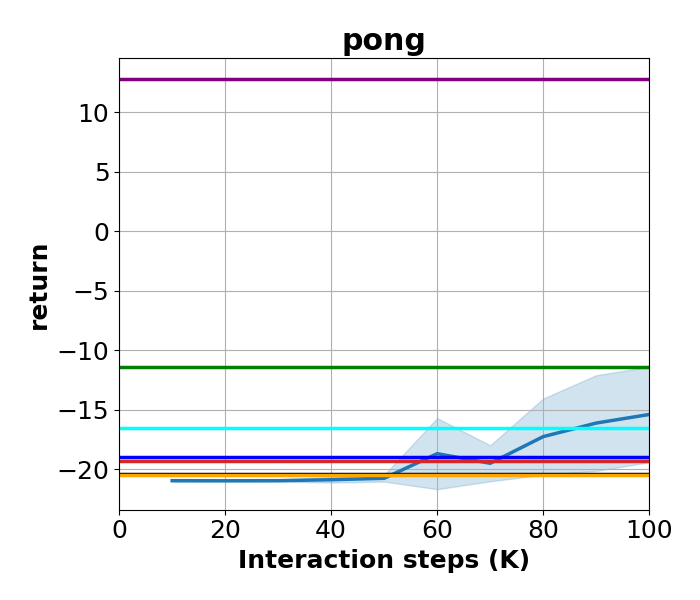}
    \includegraphics[width=0.195\textwidth]{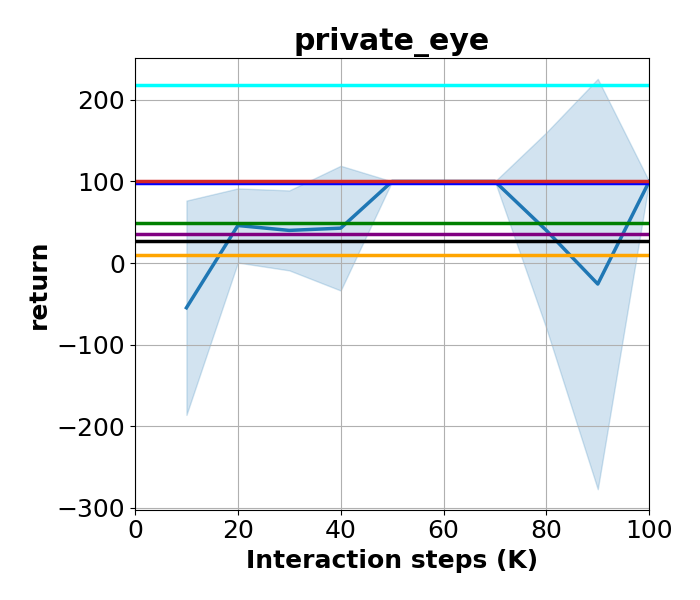}
    \includegraphics[width=0.195\textwidth]{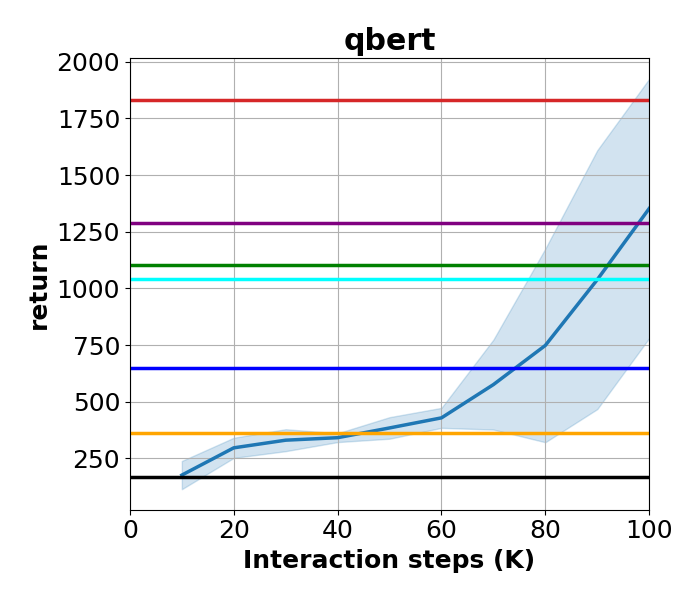}
    \includegraphics[width=0.195\textwidth]{figures/main_100k/road_runner_return.png}
    \includegraphics[width=0.195\textwidth]{figures/main_100k/seaquest_return.png}
    \includegraphics[width=0.195\textwidth]{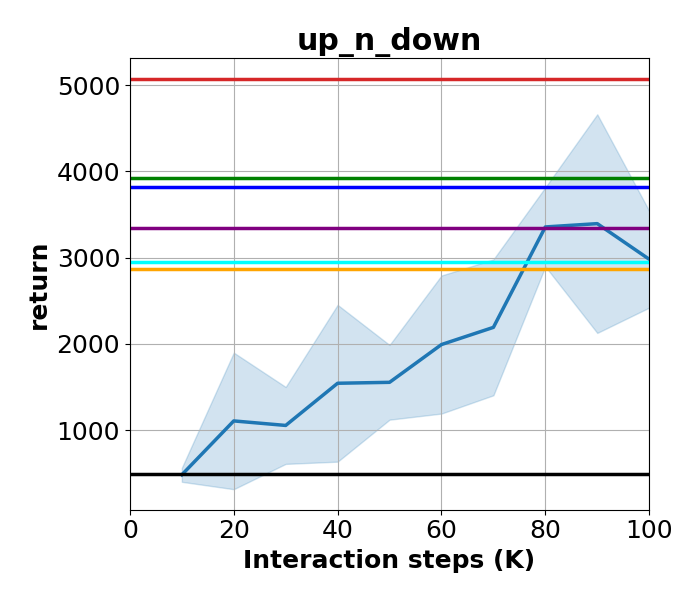}
    \caption{Comparison with baseline algorithms in 26 Atari environments at 100K interactions. The results for MeanQ show the average score and standard deviation (shaded) over 5 runs. Cited scores for baseline algorithms at 100K interactions are rendered as horizontal lines. For Random, Rainbow (“Data-Efficient”) \citep{van2019use}, and SUNRISE \citep{lee2021sunrise}, we cite the numbers reported in \citep{lee2021sunrise}. For PPO \citep{ppo}, SimPLe \citep{kaiserModelBased}, CURL \citep{laskin2020curl}, and DrQ \citep{kostrikov2021image}, we cite the numbers reported in \citet{kaiserModelBased}. } \label{atari_100k}
\end{figure*}

\begin{figure*}
    \includegraphics[width=\textwidth]{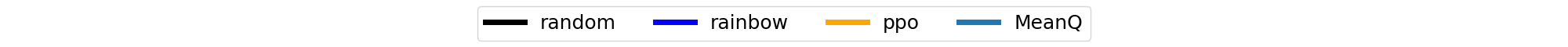}
    \includegraphics[width=0.195\textwidth]{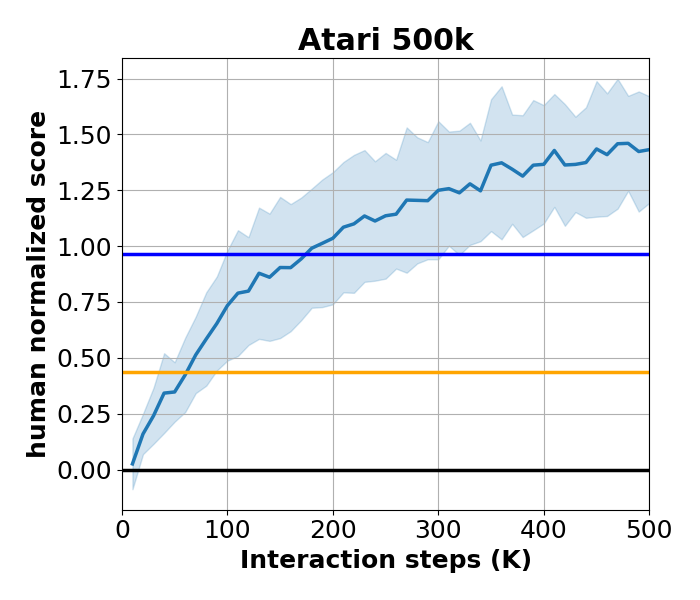}
    \includegraphics[width=0.195\textwidth]{figures/websiteplanet-dummy-700X600}
    \includegraphics[width=0.195\textwidth]{figures/websiteplanet-dummy-700X600}
    \includegraphics[width=0.195\textwidth]{figures/websiteplanet-dummy-700X600}
    \includegraphics[width=0.195\textwidth]{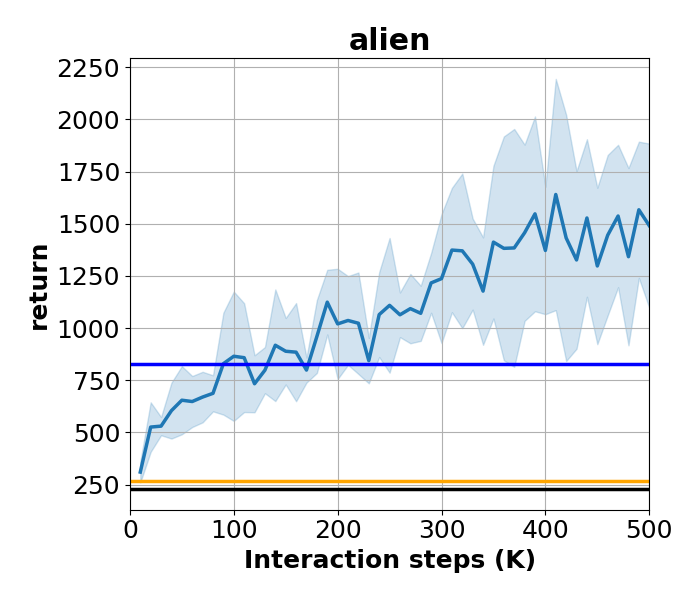}
    \includegraphics[width=0.195\textwidth]{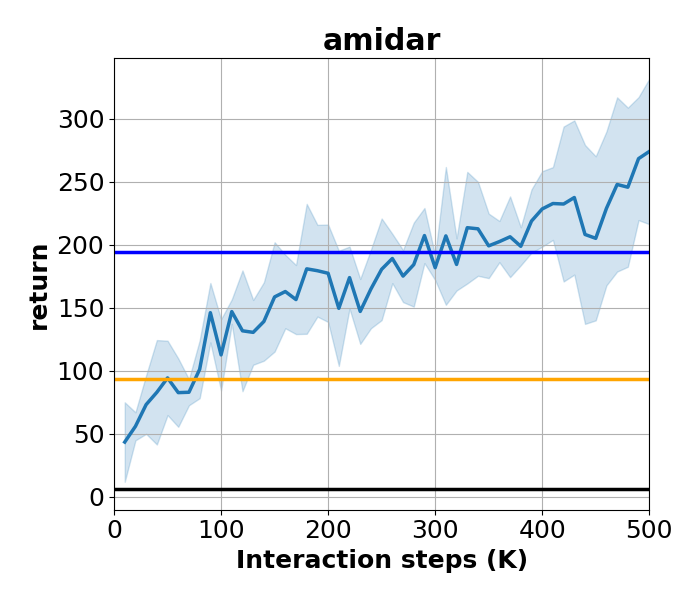}
    \includegraphics[width=0.195\textwidth]{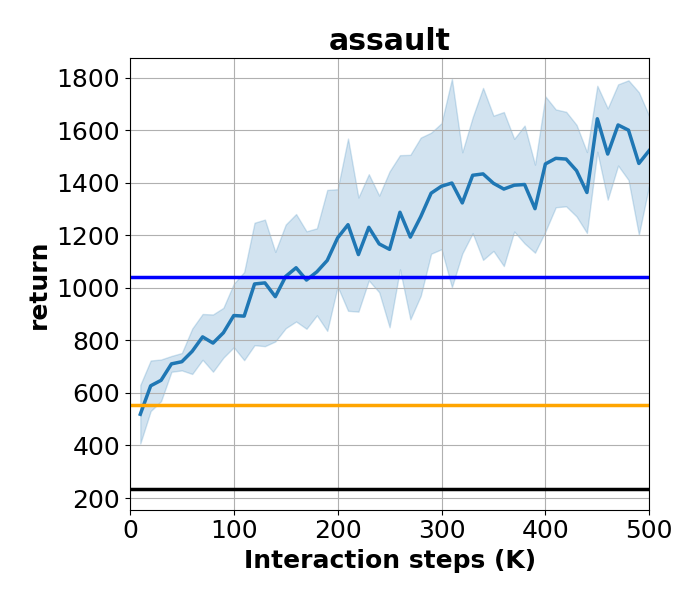}
    \includegraphics[width=0.195\textwidth]{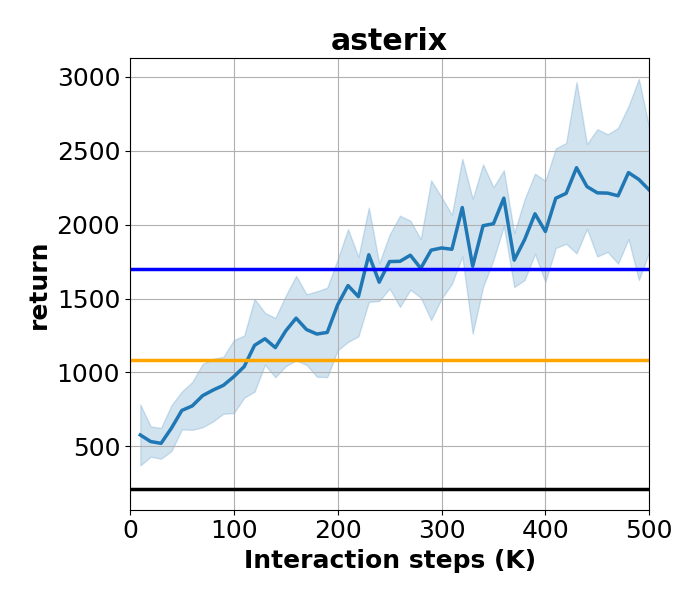}
    \includegraphics[width=0.195\textwidth]{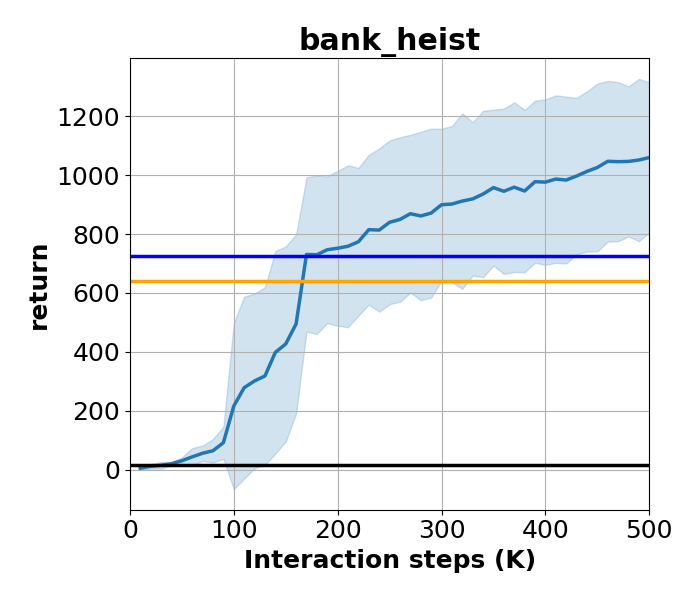}
    \includegraphics[width=0.195\textwidth]{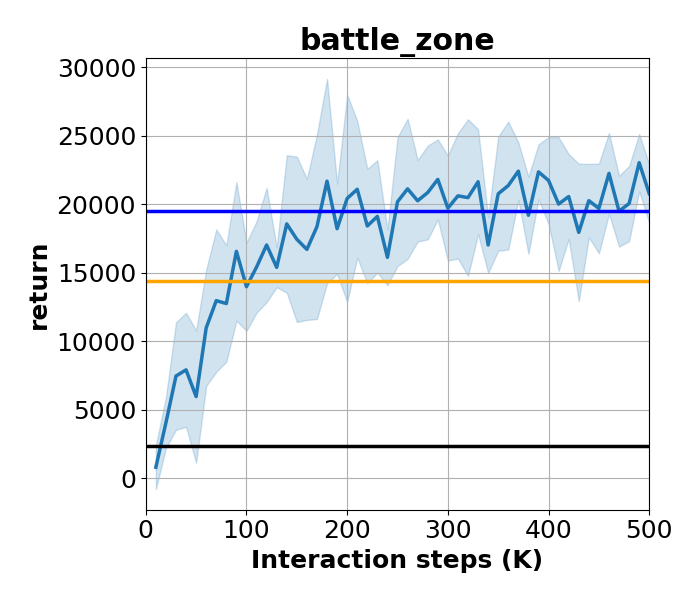}
    \includegraphics[width=0.195\textwidth]{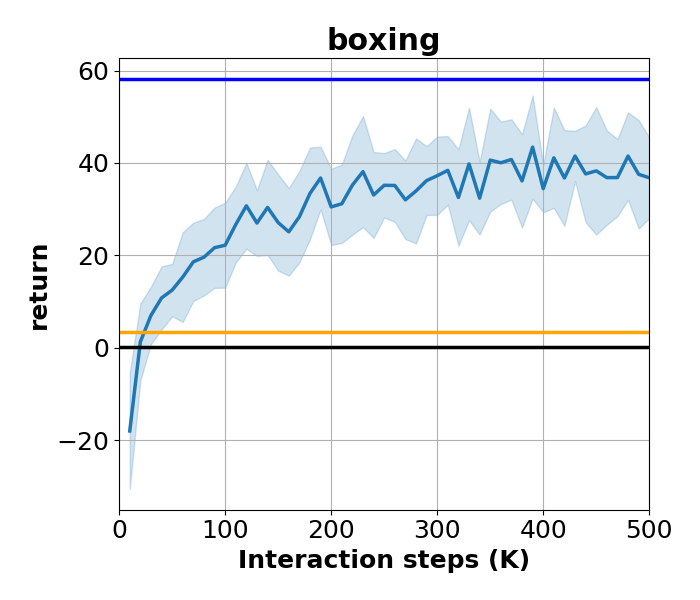}
    \includegraphics[width=0.195\textwidth]{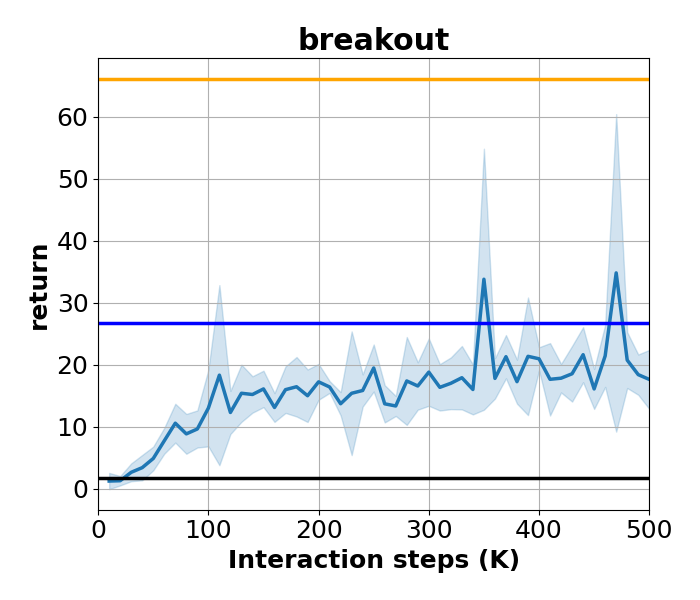}
    \includegraphics[width=0.195\textwidth]{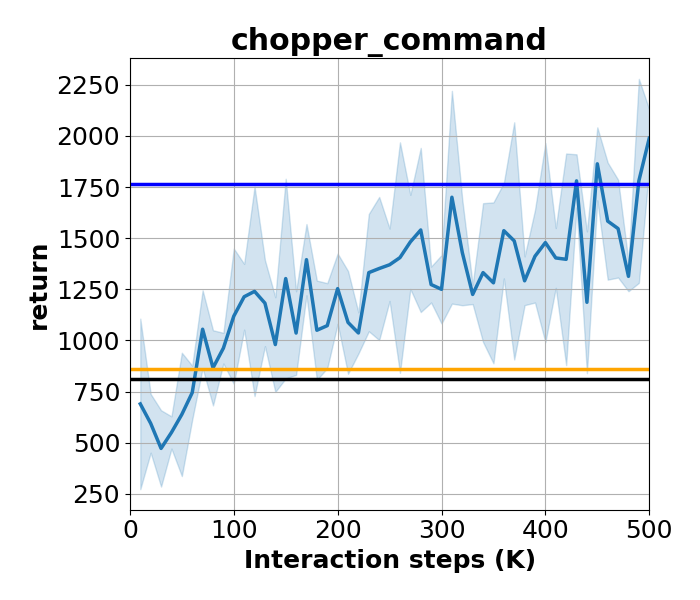}
    \includegraphics[width=0.195\textwidth]{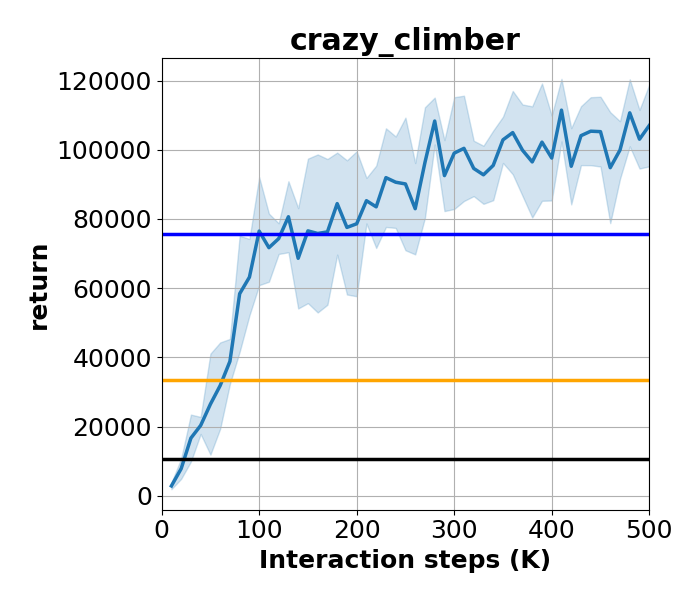}
    \includegraphics[width=0.195\textwidth]{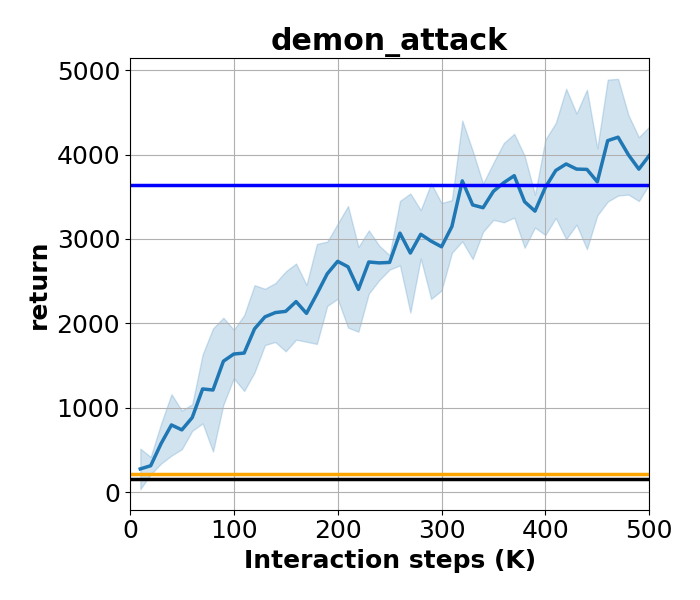}
    \includegraphics[width=0.195\textwidth]{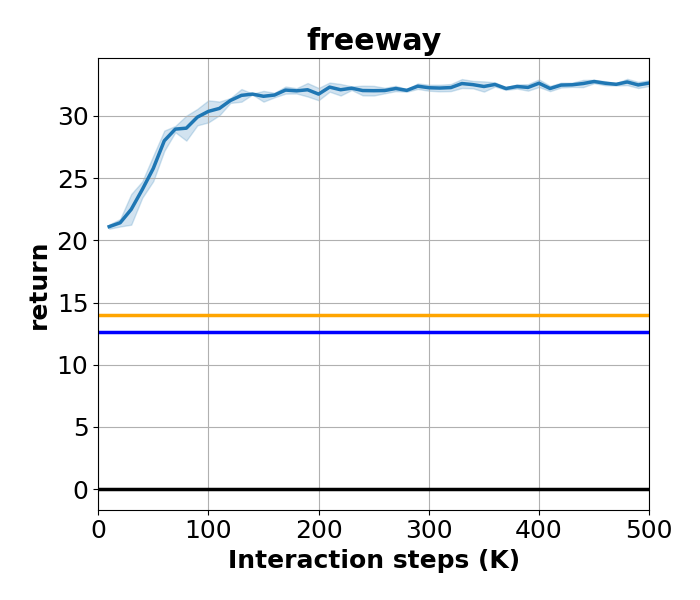}
    \includegraphics[width=0.195\textwidth]{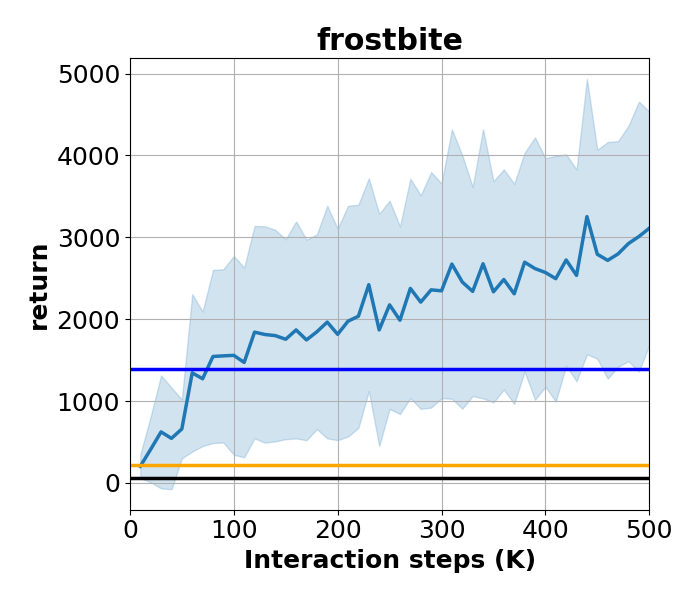}
    \includegraphics[width=0.195\textwidth]{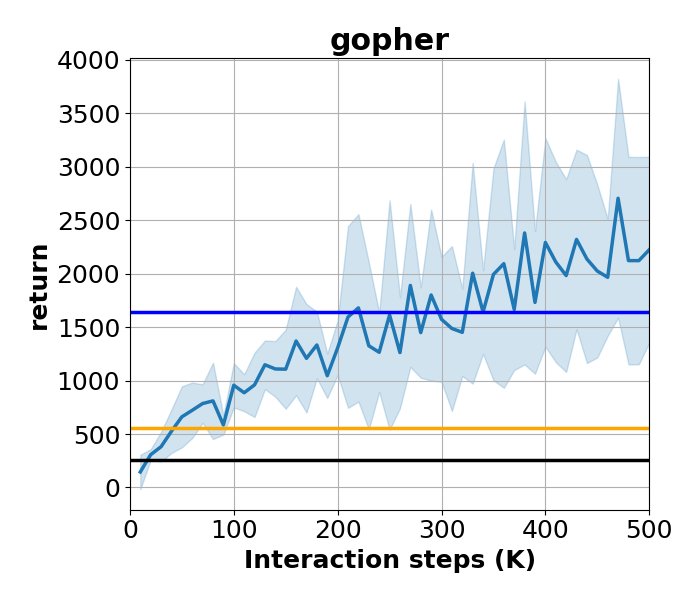}
    \includegraphics[width=0.195\textwidth]{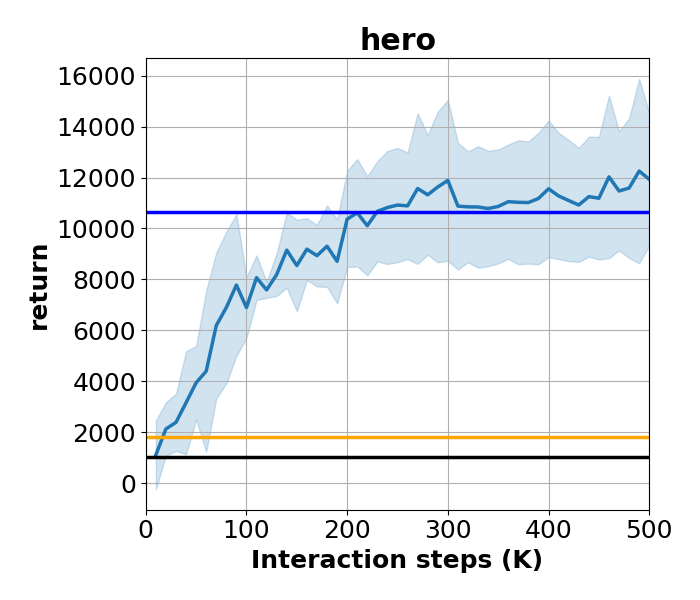}
    \includegraphics[width=0.195\textwidth]{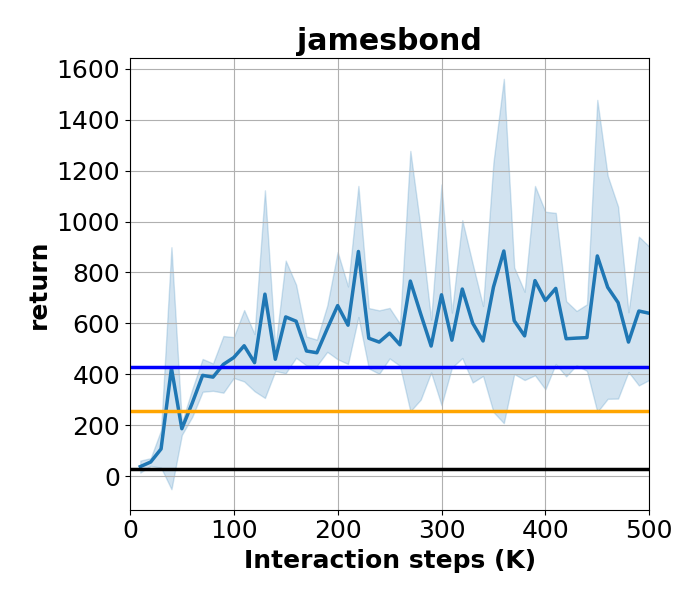}
    \includegraphics[width=0.195\textwidth]{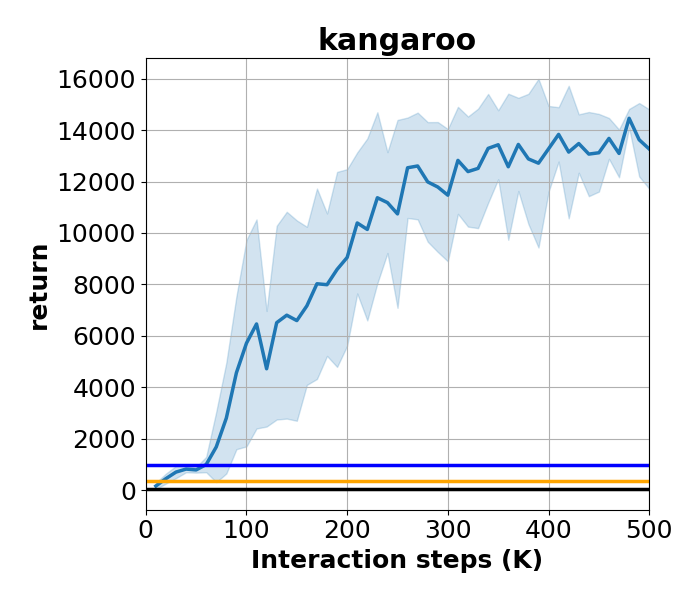}
    \includegraphics[width=0.195\textwidth]{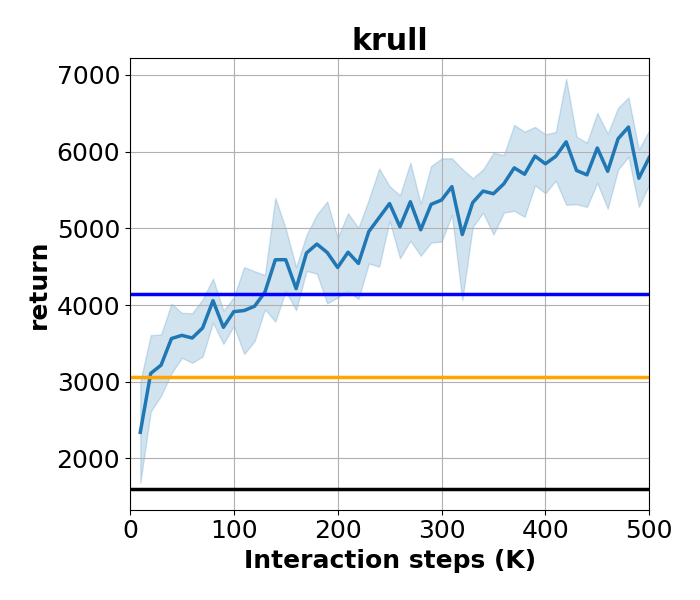}
    \includegraphics[width=0.195\textwidth]{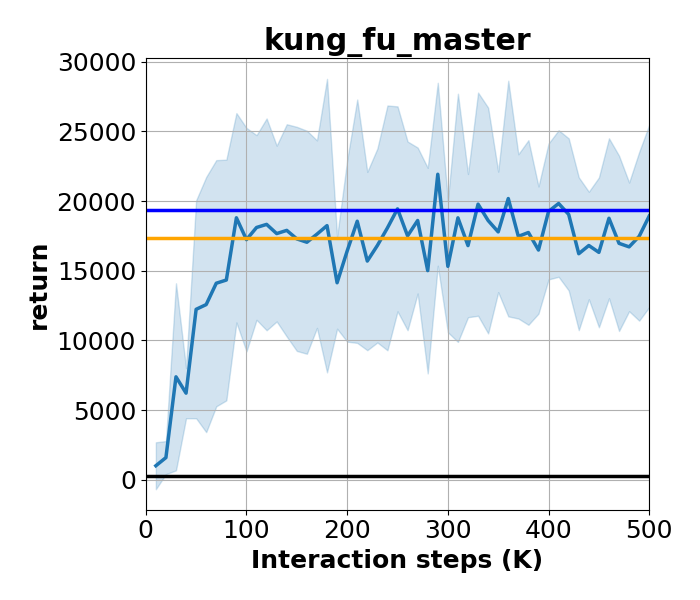}
    \includegraphics[width=0.195\textwidth]{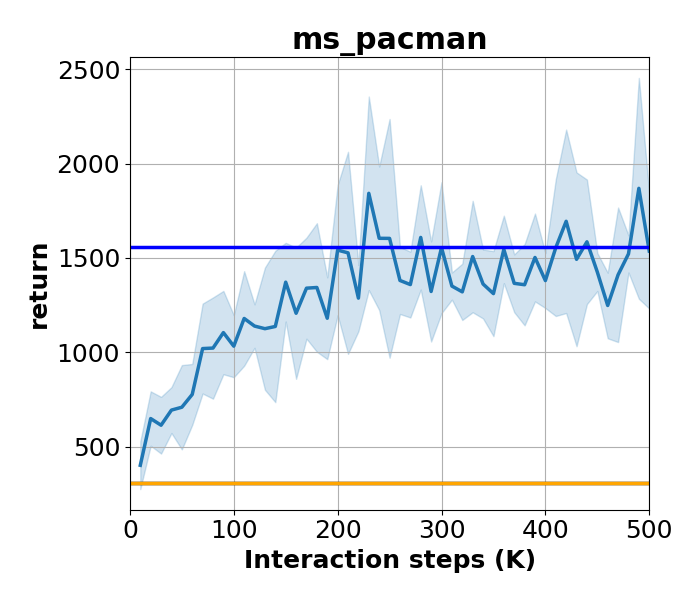}
    \includegraphics[width=0.195\textwidth]{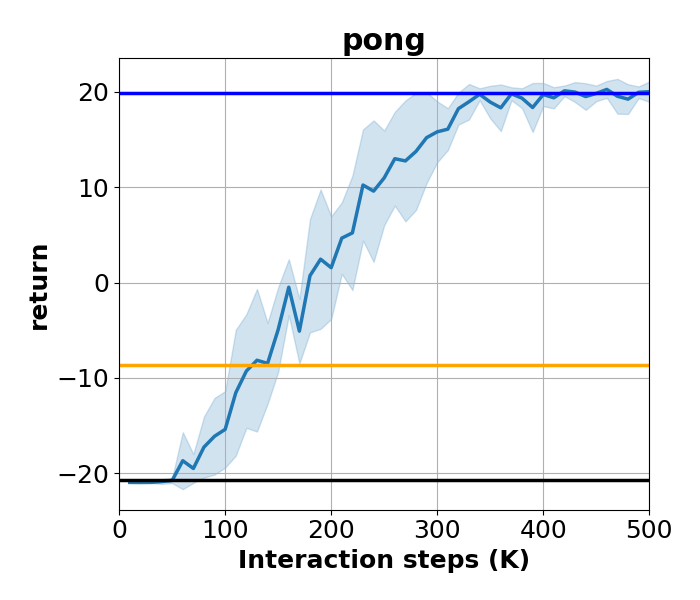}
    \includegraphics[width=0.195\textwidth]{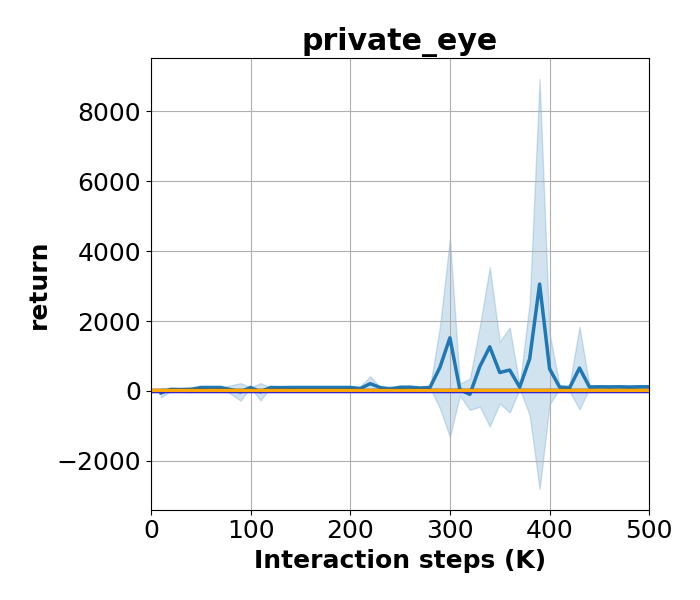}
    \includegraphics[width=0.195\textwidth]{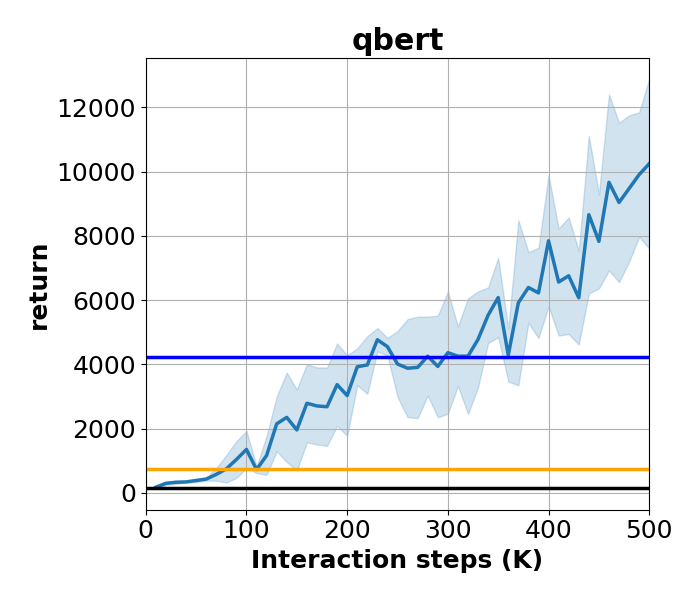}
    \includegraphics[width=0.195\textwidth]{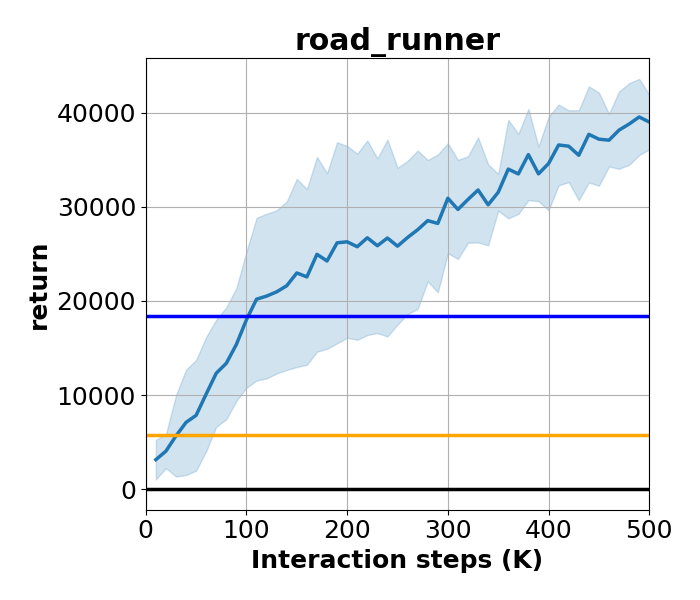}
    \includegraphics[width=0.195\textwidth]{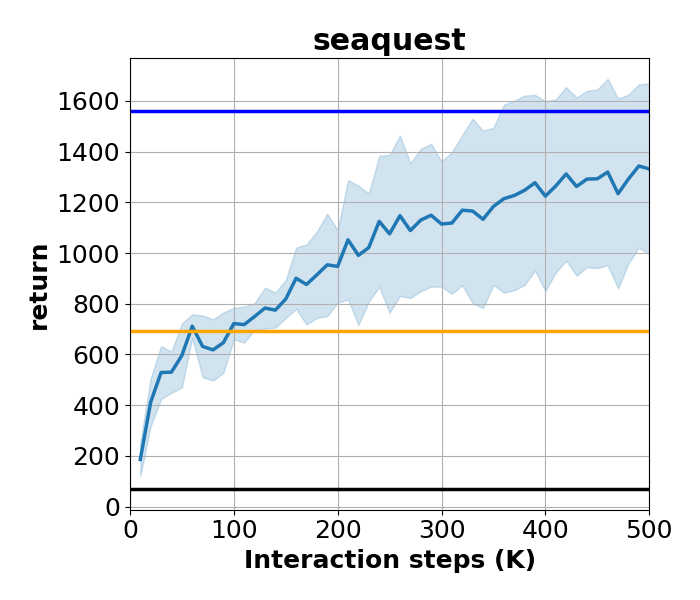}
    \includegraphics[width=0.195\textwidth]{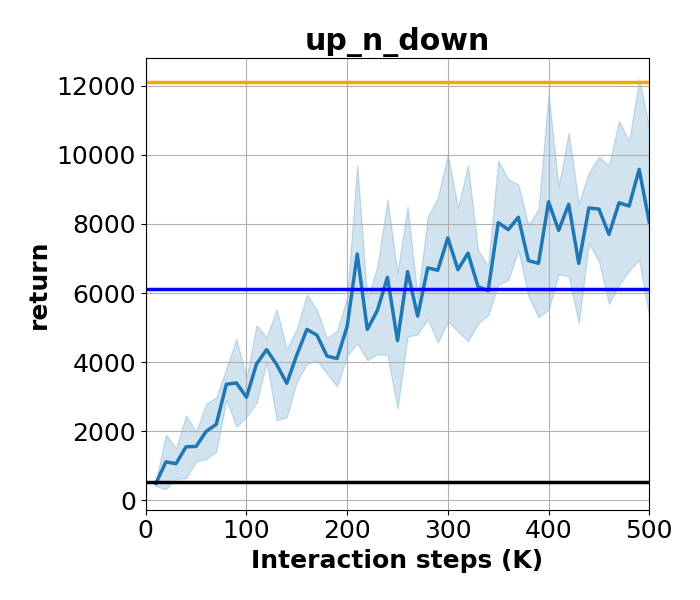}
    \caption{Comparison with baseline algorithms in 26 Atari environments at 500K interactions. The results for MeanQ show the average score and standard deviation (shaded) over 5 runs. Cited scores for baseline algorithms at 500K interactions are rendered as horizontal lines. For PPO \citep{ppo} and Rainbow (“Canonical Rainbow”) \cite{hessel2018rainbow}, we cite the numbers reported in \citet{kaiserModelBased}.}
    \label{atari_500k}
\end{figure*}

MeanQ is evaluated on the Atari Learning Environment \citep{bellemare2013arcade}. The evaluation policy is the greedy policy $\pi_\theta(s) = \argmax_a \{ \mean_k Q_{\theta_k}(s,a) \}$. Evaluation returns after the first 100K interactions are recorded in Table \ref{main_results_100k} and Figure \ref{atari_100k}. Evaluation returns after the first 500K interactions are recorded in Table \ref{main_results_500k} and Figure \ref{atari_500k}.

\newpage

\begin{table}[t] 
\caption{Performance at 100K interaction steps. The results for MeanQ show the average score and standard deviation (in parenthesis) of 5 runs. For Random, Rainbow (“Data-Efficient”), and SUNRISE, we cite the numbers reported in \citet{lee2021sunrise}. For PPO \citep{ppo}, SimPLe \citep{kaiserModelBased}, CURL \citep{laskin2020curl}, and DrQ \citep{kostrikov2021image}, we cite the numbers reported in \citet{kaiserModelBased}. Best performance for each environment is shown in bold.} \label{main_results_100k}
\fontsize{7}{8}\selectfont 
\vskip 0.15in
\begin{center}
\begin{tabular}{l|l|ll|ll|l|l|l|l|ll|l}
\toprule
 {Environment}  
&{Random} 
&{PPO} 
&{}
&{SimPLe}
&{}
&{Rainbow}
&{CURL}
&{DrQ}
&{SUNRISE}
&{MeanQ}
&{}
&{Human}
\\ \midrule
Alien               & 184.8                & 291.0                & (40.3)               & 616.9                & (252.2)              & 789.0                & 558.2                & 761.4                & \textbf{872.0}       & 864.8                & (309.7)              & 7128.0               \\
Amidar              & 11.8                 & 56.5                 & (20.8)               & 74.3                 & (28.3)               & 118.5                & \textbf{142.1}       & 97.3                 & 122.6                & 112.5                & (28.3)               & 1720.0               \\
Assault             & 233.7                & 424.2                & (55.8)               & 527.2                & (112.3)              & 413.0                & 600.6                & 489.1                & 594.8                & \textbf{894.4}       & (120.7)              & 742.0                \\
Asterix             & 248.8                & 385.0                & (104.4)              & \textbf{1128.3}      & (211.8)              & 533.3                & 734.5                & 637.5                & 755.0                & 971.5                & (246.7)              & 8503.0               \\
BankHeist           & 15.0                 & 16.0                 & (12.4)               & 34.2                 & (29.2)               & 97.7                 & 131.6                & 196.6                & \textbf{266.7}       & 215.6                & (282.4)              & 753.0                \\
BattleZone          & 2895.0               & 5300.0               & (3655.1)             & 4031.2               & (1156.1)             & 7833.3               & 14870.0              & 13520.6              & \textbf{15700.0}     & 13980.0              & (3225.0)             & 37188.0              \\
Boxing              & 0.3                  & -3.9                 & (6.4)                & 7.8                  & (10.1)               & 0.6                  & 1.2                  & 6.9                  & 6.7                  & \textbf{22.2}        & (9.2)                & 12.0                 \\
Breakout            & 0.9                  & 5.9                  & (3.3)                & \textbf{16.4}        & (6.2)                & 2.3                  & 4.9                  & 14.5                 & 1.8                  & 13.0                 & (6.2)                & 30.0                 \\
ChopperCommand      & 671.0                & 730.0                & (199.0)              & 979.4                & (172.7)              & 590.0                & 1058.5               & 646.6                & 1040.0               & \textbf{1118.8}      & (330.5)              & 7388.0               \\
CrazyClimber        & 7339.5               & 18400.0              & (5275.1)             & 62583.6              & (16856.8)            & 25426.7              & 12146.5              & 19694.1              & 22230.0              & \textbf{76490.0}     & (15610.2)            & 35829.0              \\
DemonAttack         & 140.0                & 192.5                & (83.1)               & 208.1                & (56.8)               & 688.2                & 817.6                & 1222.2               & 919.8                & \textbf{1636.8}      & (287.6)              & 1971.0               \\
Freeway             & 0.0                  & 8.0                  & (9.8)                & 16.7                 & (15.7)               & 28.7                 & 26.7                 & 15.4                 & 30.2                 & \textbf{30.4}        & (0.9)                & 30.0                 \\
Frostbite           & 74.0                 & 214.0                & (10.2)               & 65.2                 & (31.5)               & 1478.3               & 1181.3               & 449.7                & \textbf{2026.7}      & 1557.8               & (1214.6)             & 4334.7               \\
Gopher              & 245.9                & 246.0                & (103.3)              & 596.8                & (183.5)              & 348.7                & 669.3                & 598.4                & 654.7                & \textbf{956.0}       & (207.5)              & 2412.0               \\
Hero                & 224.6                & 569.0                & (1100.9)             & 2656.6               & (483.1)              & 3675.7               & 6279.3               & 4001.6               & \textbf{8072.5}      & 6893.2               & (1192.8)             & 30826.0              \\
Jamesbond           & 29.2                 & 65.0                 & (46.4)               & 100.5                & (36.8)               & 300.0                & \textbf{471.0}       & 272.3                & 390.0                & 466.2                & (80.1)               & 303.0                \\
Kangaroo            & 42.0                 & 140.0                & (102.0)              & 51.2                 & (17.8)               & 1060.0               & 872.5                & 1052.4               & 2000.0               & \textbf{5714.0}      & (4012.9)             & 3035.0               \\
Krull               & 1543.3               & 3750.4               & (3071.9)             & 2204.8               & (776.5)              & 2592.1               & \textbf{4229.6}      & 4002.3               & 3087.2               & 3913.4               & (185.7)              & 2666.0               \\
KungFuMaster        & 616.5                & 4820.0               & (983.2)              & 14862.5              & (4031.6)             & 8600.0               & 14307.8              & 7106.4               & 10306.7              & \textbf{17232.0}     & (8019.1)             & 22736.0              \\
MsPacman            & 235.2                & 496.0                & (379.8)              & 1480.0               & (288.2)              & 1118.7               & 1465.5               & 1065.6               & \textbf{1482.3}      & 1032.0               & (164.5)              & 6952.0               \\
Pong                & -20.4                & -20.5                & (0.6)                & \textbf{12.8}        & (17.2)               & -19.0                & -16.5                & -11.4                & -19.3                & -15.4                & (4.0)                & 15.0                 \\
PrivateEye          & 26.6                 & 10.0                 & (20.0)               & 35.0                 & (60.2)               & 97.8                 & \textbf{218.4}       & 49.2                 & 100.0                & 100.0                & (0.0)                & 69571.0              \\
Qbert               & 166.1                & 362.5                & (117.8)              & 1288.8               & (1677.9)             & 646.7                & 1042.4               & 1100.9               & \textbf{1830.8}      & 1352.2               & (571.9)              & 13455.0              \\
RoadRunner          & 0.0                  & 1430.0               & (760.0)              & 5640.6               & (3936.6)             & 9923.3               & 5661.0               & 8069.8               & 11913.3              & \textbf{18012.5}     & (7229.0)             & 7845.0               \\
Seaquest            & 61.1                 & 370.0                & (103.3)              & 683.3                & (171.2)              & 396.0                & 384.5                & 321.8                & 570.7                & \textbf{721.8}       & (62.4)               & 42055.0              \\
UpNDown             & 488.4                & 2874.0               & (1105.8)             & 3350.3               & (3540.0)             & 3816.0               & 2955.2               & 3924.9               & \textbf{5074.0}      & 2982.5               & (560.9)              & 11693.0              \\
\bottomrule
\end{tabular}
\end{center}
\end{table}

\begin{table}[b] 
\caption{Performance at 500K interaction steps. The results for MeanQ show the average score and standard deviation (in parenthesis) of 5 runs. For PPO \citep{ppo} and Rainbow (“Canonical Rainbow”) \cite{hessel2018rainbow} we cite the numbers reported in \citet{kaiserModelBased}. Best performance for each environment is shown in bold.}\label{main_results_500k}
\fontsize{7}{8}\selectfont 
\vskip 0.15in
\begin{center}
\begin{tabular}{l|l|ll|ll|ll|l}
\toprule
 {Environment}  
&{Random} 
&{PPO} 
&{}
&{Rainbow}
&{}
&{MeanQ}
&{}
&{Human}
\\ \midrule
Alien               & 227.8                & 269.0                & (203.4)              & 828.6                & (54.2)               & \textbf{1490.7}      & (393.1)              & 7128.0               \\
Amidar              & 5.8                  & 93.2                 & (36.7)               & 194.0                & (34.9)               & \textbf{274.1}       & (57.7)               & 1720.0               \\
Assault             & 233.7                & 552.3                & (110.4)              & 1041.5               & (92.1)               & \textbf{1522.4}      & (131.1)              & 742.0                \\
Asterix             & 210.0                & 1085.0               & (354.8)              & 1702.7               & (162.8)              & \textbf{2234.5}      & (419.2)              & 8503.0               \\
BankHeist           & 14.2                 & 641.0                & (352.8)              & 727.3                & (198.3)              & \textbf{1060.2}      & (255.0)              & 753.0                \\
BattleZone          & 2360.0               & 14400.0              & (6476.1)             & 19507.1              & (3193.3)             & \textbf{20770.0}     & (2127.6)             & 37188.0              \\
Boxing              & 0.1                  & 3.5                  & (3.5)                & \textbf{58.2}        & (16.5)               & 36.8                 & (8.7)                & 12.0                 \\
Breakout            & 1.7                  & \textbf{66.1}        & (114.3)              & 26.7                 & (2.4)                & 14.9                 & (5.6)                & 30.0                 \\
ChopperCommand      & 811.0                & 860.0                & (285.3)              & 1765.2               & (280.7)              & \textbf{1816.2}      & (322.0)              & 7388.0               \\
CrazyClimber        & 10780.5              & 33420.0              & (3628.3)             & 75655.1              & (9439.6)             & \textbf{107003.0}    & (11795.9)            & 35829.0              \\
DemonAttack         & 152.1                & 216.5                & (96.2)               & 3642.1               & (478.2)              & \textbf{3988.2}      & (341.7)              & 1971.0               \\
Freeway             & 0.0                  & 14.0                 & (9.8)                & 12.6                 & (15.4)               & \textbf{32.6}        & (0.2)                & 30.0                 \\
Frostbite           & 65.2                 & 214.0                & (10.2)               & 1386.1               & (321.7)              & \textbf{3110.0}      & (1426.4)             & 4334.7               \\
Gopher              & 257.6                & 560.0                & (118.8)              & 1640.5               & (105.6)              & \textbf{2223.4}      & (869.4)              & 2412.0               \\
Hero                & 1027.0               & 1824.0               & (1461.2)             & 10664.3              & (1060.5)             & \textbf{11928.8}     & (2596.1)             & 30826.0              \\
Jamesbond           & 29.0                 & 255.0                & (101.7)              & 429.7                & (27.9)               & \textbf{640.0}       & (261.8)              & 303.0                \\
Kangaroo            & 52.0                 & 340.0                & (407.9)              & 970.9                & (501.9)              & \textbf{13263.0}     & (1542.6)             & 3035.0               \\
Krull               & 1598.0               & 3056.1               & (1155.5)             & 4139.4               & (336.2)              & \textbf{5923.9}      & (352.7)              & 2666.0               \\
KungFuMaster        & 258.5                & 17370.0              & (10707.6)            & \textbf{19346.1}     & (3274.4)             & 18932.0              & (6543.1)             & 22736.0              \\
MsPacman            & 307.3                & 306.0                & (70.2)               & \textbf{1558.0}      & (248.9)              & 1536.0               & (304.5)              & 6952.0               \\
Pong                & -20.7                & -8.6                 & (14.9)               & 19.9                 & (0.4)                & \textbf{20.0}        & (1.1)                & 15.0                 \\
PrivateEye          & 24.9                 & 20.0                 & (40.0)               & -6.2                 & (89.8)               & \textbf{120.0}       & (40.0)               & 69571.0              \\
Qbert               & 163.9                & 757.5                & (78.9)               & 4241.7               & (193.1)              & \textbf{10245.5}     & (2640.8)             & 13455.0              \\
RoadRunner          & 11.5                 & 5750.0               & (5259.9)             & 18415.4              & (5280.0)             & \textbf{38992.5}     & (2890.7)             & 7845.0               \\
Seaquest            & 68.4                 & 692.0                & (48.3)               & \textbf{1558.7}      & (221.2)              & 1331.8               & (337.5)              & 42055.0              \\
UpNDown             & 533.4                & \textbf{12126.0}     & (1389.5)             & 6120.7               & (356.8)              & 8051.6               & (2730.7)             & 11693.0              \\
\bottomrule
\end{tabular}
\end{center}

\end{table}

\newpage

\end{document}